\documentclass[10pt,journal,compsoc]{IEEEtran}

\ifCLASSOPTIONcompsoc
  \usepackage[nocompress]{cite}
\else
  \usepackage{cite}
\fi

\usepackage[utf8]{inputenc}
\usepackage[T1]{fontenc}
\usepackage{xcolor,soul}
\usepackage{graphicx}%
\usepackage{multirow}%
\usepackage{amsmath,amssymb,amsfonts}%
\usepackage{amsthm}%
\usepackage{mathrsfs}%
\usepackage{xcolor}%
\usepackage{pifont}
\usepackage{diagbox}
\usepackage{listings}%
\usepackage{tabularx}
\usepackage{xr}
\usepackage{hyperref}
\usepackage{booktabs}%
\usepackage{subfigure}

\def \OURS{MolMark}

\begin{document}

\title{MolMark: Safeguarding Molecular Structures through Learnable Atom-Level Watermarking}

\author{Runwen Hu, Peilin Chen, Keyan Ding, and Shiqi Wang*, \emph{Senior Member, IEEE}

\IEEEcompsocitemizethanks{
\IEEEcompsocthanksitem Runwen Hu, Peilin Chen, and Shiqi Wang are with the Department of Computer Science, City University of Hong Kong, Hong Kong (e-mail: runwenhu@cityu.edu.hk; plchen3@cityu.edu.hk; shiqwang@cityu.edu.hk)
\IEEEcompsocthanksitem Keyan Ding is with the ZJU-Hangzhou Global Scientific and Technological Innovation Center, Zhejiang University, Hangzhou, China (e-mail: dingkeyan@zju.edu.cn;)
\IEEEcompsocthanksitem Corresponding author: Shiqi Wang.
}
}

\markboth{Submitted to IEEE Transactions on Pattern Analysis and Machine Intelligence}{Hu et al: MolMark: Safeguarding Molecular Structures with Atom-Level Watermarking}

\IEEEtitleabstractindextext{%
\begin{abstract}
AI-driven molecular generation is reshaping drug discovery and materials design, yet the lack of protection mechanisms leaves AI-generated molecules vulnerable to unauthorized reuse and provenance ambiguity. Such limitation undermines both scientific reproducibility and intellectual property security. To address this challenge, we propose the first deep learning based watermarking framework for molecules (\OURS{}), which is exquisitely designed to embed high-fidelity digital signatures into molecules without compromising molecular functionalities. \OURS{} learns to modulate the chemically meaningful atom-level representations and enforce geometric robustness through SE(3)-invariant features, maintaining robustness under rotation, translation, and reflection. Additionally, \OURS{} integrates seamlessly with AI-based molecular generative models, enabling watermarking to be treated as a learned transformation with minimal interference to molecular structures. Experiments on benchmark datasets (QM9, GEOM-DRUG) and state-of-the-art molecular generative models (GeoBFN, GeoLDM) demonstrate that \OURS{} can embed 16-bit watermarks while retaining more than 90\% of essential molecular properties, preserving downstream performance, and enabling >95\% extraction accuracy under SE(3) transformations. \OURS{} establishes a principled pathway for unifying molecular generation with verifiable authorship, supporting trustworthy and accountable AI-driven molecular discovery.
\end{abstract}

}

\maketitle

\IEEEdisplaynontitleabstractindextext
\IEEEpeerreviewmaketitle

\begin{figure*}[htbp]
  \centering
  \includegraphics[width=0.8\textwidth]{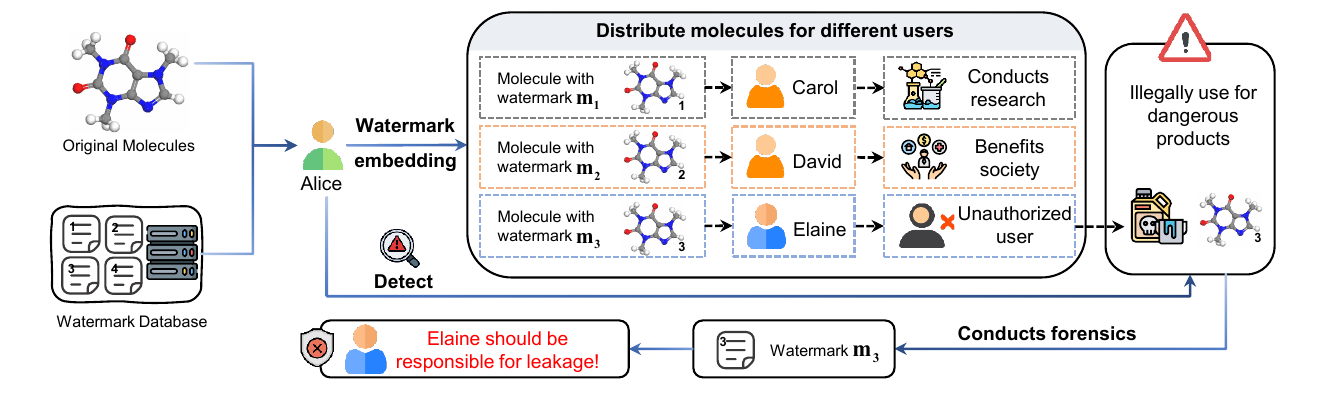}
  \caption{Application scenarios of \OURS{} in protecting molecules to tracking data leakage. Alice applies \OURS{} to embed watermarks into molecules and distributes uniquely watermarked molecules to different users. When Elaine leaks her copy to unauthorized users, Alice can detect the leakage and successfully trace it back to Elaine.}
  \label{Fig1}
\end{figure*}

\IEEEraisesectionheading{\section{Introduction}}
\label{section:Introduction}

Synthetic molecules are foundational to advances in biology, chemistry, and medicine, enabling breakthroughs ranging from protein engineering~\cite{yang2019machine,luo2021ecnet} to drug discovery~\cite{pandey2022transformational,yu2024deep}. Traditional molecular designs rely on labor-intensive and trial-and-error workflows, limited by human intuition and experimental throughput. Recently, the convergence of artificial intelligence (AI) with molecular sciences reshapes the paradigms, enabling rapid and data-driven discovery of functional compounds with precisely tailored properties~\cite{de2019synthetic,segler2018planning,shafiq2025generative,goswami2023artificial}. In particular, the emergence of 3D molecular generative models, such as autoregressive frameworks~\cite{gebauer2019symmetry}, flow-based methods~\cite{luo2022autoregressive}, diffusion models~\cite{xu2023geometric,anand2022protein,peng2023moldiff}, and Bayesian Flow Networks~\cite{graves2023bayesian,song2023unified,qu2024molcraft,atkinson2025protein,guloglu2025abbfn2}, have dramatically expanded the scope and fidelity of molecular generation, accelerating AI-driven discovery of molecules.

Despite these advances, a critical challenge remains unaddressed: the lack of reliable mechanism to protect the intellectual property (IP) of AI-generated molecules~\cite{wang2025call,fan2025safeprotein,feldman2025resilient}. In contrast to other AI-generated digital assets~\cite{li2022diffusion,liu2024sora,cotton2024chatting}, such as images or text, molecular structures occupy a regulatory and technical gray zone. Molecular structures, as functional and naturally occurring entities, are not eligible for copyright protection. Although patents can cover novel compounds, the process is costly and typically unsuitable for early-stage in silico exploration. Consequently, AI-generated molecular assets are highly vulnerable to unauthorized appropriation, redistribution, and false attribution. As shown in Fig~\ref{Fig1}, malicious actors may claim ownership of shared molecules, redistribute molecules without credit or introduce modifications to evade detection. Such risks are amplified in high-stakes domains such as pharmaceuticals, where provenance influences safety, liability, and regulatory approval~\cite{pokharel2025street}. Tracing the origin of molecules used in preclinical or clinical studies becomes infeasible. Therefore, the absence of IP safeguards may further discourage the open sharing of generative models and datasets, constraining collaboration and slowing scientific progress.

Digital watermarking methods offers a compelling technical solution to this challenge. By imperceptibly embedding a unique signature into the data, watermarking methods enable ownership verification, provenance tracking, and tamper detection. Currently, watermarking methods are well established for multimedia domains~\cite{zhu2018hidden,liu2019novel,jia2021mbrs,zhang2021deep,fang2022end,huang2023arwgan,zhang2024robust} and macromolecular structures like proteins~\cite{zhang2024foldmark,chen2025enhancing} and DNA~\cite{zhang2025securing,zhang2025genebreaker}, yet the applications to AI-generated molecules remains unexplored. Directly applying previous watermarking methods for molecular protection raise significant difficulties, which stem from three inherent challenges of molecular systems. First, molecular structures are compact and atomically sparse, leaving limited flexibility for watermark embedding. Second, molecules exhibit extreme sensitivity to perturbations, in which small deviations in bond lengths or angles may break chemical bonds, destabilize conformations, or abolish biological activity. Third, molecular structures and function are preserved after processing by SE(3) transformations (i.e., rotations, translations, and reflections), indicating watermarking methods must exhibits strong robustness. However, existing watermarking methods cannot solve these challenges, promoting the research of developing new methods for molecular protection.

To address these challenges, we propose \OURS{}, the first deep learning-based watermarking method tailored for AI-generated molecular structures. \OURS{} learns to embed watermarks at atom level by modulating chemically informed features in a manner that preserves molecular stability and functionality. Crucially, the embedding and extraction processes operate on SE(3)-invariant representations, ensuring robustness to geometric transformations. To further enhance reliability, we propose a dynamically balanced training strategy, which mitigates numerical instability and improves overall performance. We rigorously evaluate \OURS{} on standard benchmarks (QM9 and GEOM-DRUG) and state-of-the-art generative models (GeoBFN and GeoLDM). Our results demonstrate that \OURS{} can embed 16-bit watermarks with minimal impact, including maintaining molecular stability above 94.6\%, atomic stability exceeds 97.6\%, and downstream docking performance virtually unchanged. Moreover, watermark extraction accuracy surpasses 95\% under SE(3) transformations. By achieving IP protection and maintaining scientific utility of molecules, \OURS{} establishes a foundational framework for responsible and trustworthy AI-driven molecular design. The main contributions are summarized as follows:

\begin{itemize}
\item \textbf{We propose the first deep learning-based watermarking method for AI-generated molecules.} \OURS{} brings digital watermarking from traditional multimedia and protein domains into the atomic and molecular systems, enabling reliable signatures to be embedded within small and geometrically fragile molecules while preserving chemical stability and functional integrity.
\item \textbf{We develop SE(3)-invariant features for embedding and extraction pipeline.} By ensuring robustness to rotations, translations, and reflections, \OURS{} achieve watermark extraction that remains stable across coordinate transformations and geometric perturbations, extending geometric invariance from molecular generation to molecular IP protection.
\item \textbf{We propose a dynamically balanced training strategy that stabilizes the trade-off between molecular properties and watermark extractability.} The optimization scheme mitigates numerical instability and harmonizes robustness and molecular fidelity, enabling \OURS{} to achieve high structural stability and watermark extraction accuracy.
\end{itemize}

The rest of this paper is organized as follows. Section~\ref{section:Proposed Method} introduces the proposed method \OURS{} in detail. In Section~\ref{section:Experimental Results}, we conduct exhaustive experiments and comparisons to evaluate the performance of \OURS{}. Finally, we conclude this paper in Section~\ref{section:Conclusion}.

\begin{figure*}[htbp]
  \centering
  \includegraphics[width=0.8\textwidth]{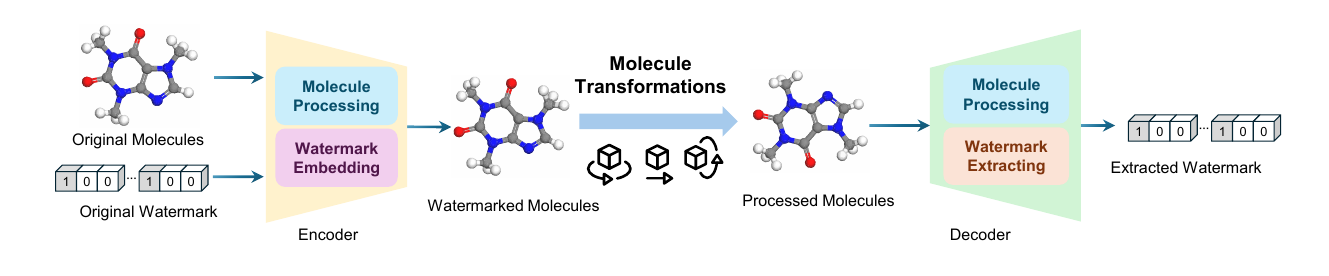}
  \caption{The framework of \OURS{}. The encoder $\mathcal{E}_\phi$ embeds watermarks into original molecules, generating the watermarked molecules with minimal impact on the molecular properties and functionalities. Molecular transformations are applied to simulate the real-world process on molecules. The decoder $\mathcal{D}_\theta$ effectively extracts the watermarks from the watermarked molecules, enabling reliable copyright protection.}
  \label{Fig2}
\end{figure*}

\section{Proposed Method}
\label{section:Proposed Method}

In this section, we propose a new watermarking method \OURS{} specifically designed for molecular protection. Unlike conventional multimedia data, molecules are composed of complex and multimodal data, including discrete atomic composition and continuous three-dimensional spatial arrangement. Table~\ref{Table1} compares the structural characteristics of images, proteins, and molecules. Images consist of discrete pixels with high redundancy and low sensitivity to pixel modifications. Proteins exhibit structural complexity, but include non-critical regions and moderate tolerance to modifications. In contrast, molecules are governed by strict chemical constraints and highly sensitive to minimal alterations. To effectively protect molecules, \OURS{} is exquisitely designed to achieve a practical balance among fidelity, imperceptibility, and robustness through deep learning based techniques.

Fig.~\ref{Fig2} shows the framework of \OURS{}, which consists of a watermark encoder $\mathcal{E}_\phi$ and a decoder $\mathcal{D}_\theta$. The encoder learns to embed watermarks into molecular structures while preserving molecular properties. After that, the resulting watermarked molecules are subjected to common spatial transformations (e.g., rotation, translation, and reflection), simulating typical post-processing distortions encountered in practice. Finally, the decoder extracts the embedded watermarks from transformed molecules, enabling reliable copyright protection.

\begin{table*}[htbp]
\caption{Comparative analysis of the difference between images, proteins, and molecules.}
\centering
\begin{tabular}{lccc}
\toprule
                       & Images                                    & Proteins                               & Molecules                                     \\ \midrule
Dimension              & 2D/3D                                     & 3D                                     & 3D             \\
Structure              & Discrete pixels                           & Multimodal data                        & Multimodal data          \\
Property               & High redundancy                           & Non-critical regions                   & Strict chemical constraints \\
Sensitivity            & Low                                       & Moderate                               & High                                     \\
Space                  & High (128$\times$128$\times$3 pixels)                   & Moderate (500$\times$37 atoms)                & Extremely low (200 atoms)  \\ \bottomrule
\end{tabular}
\label{Table1}
\end{table*}

\subsection{Watermark Embedding}
\label{section:Watermark Embedding}

The encoder $\mathcal{E}_\phi$ is designed to embed watermarks into molecules while preserving the intrinsic properties. Let $d$ denote the dimension of node features and $N$ the number of atoms in a molecule. Each 3D molecule can be represented as the geometry $\mathcal{G}=\left\langle {\mathbf{p},\mathbf{h}} \right\rangle  \in \mathbb{R}^{N \times (3+d)}$, where $\mathbf{p} = (\mathbf{p}_1, ..., \mathbf{p}_N) \in \mathbb{R}^{N \times 3}$ denotes the atom positions and $\mathbf{h} = (\mathbf{h}_1, ..., \mathbf{h}_N) \in \mathbb{R}^{N \times d}$ represents the node features, which are composed of the atom type $\mathbf{t} = (\mathbf{t}_1, ..., \mathbf{t}_N) \in \mathbb{R}^{N \times e}$, the atom charge $\mathbf{c} = (\mathbf{c}_1, ..., \mathbf{c}_N) \in \mathbb{R}^{N \times 1}$, and the edge indexes $\mathbf{e} = (\mathbf{e}_1, ..., \mathbf{e}_N) \in \mathbb{R}^{N \times (N-1)}$. Specifically, atom types are represented as one-hot vectors, where $e$ is the number of atomic species in the dataset (e.g., $e=5$ for QM9 dataset and $e=16$ for GEOM-DRUG dataset).

During embedding, a binary watermark $\mathbf{m} = (\mathbf{m}_1, \ldots, \mathbf{m}_L) \in \{0,1\}^{1 \times L}$ with length $L$ is embedded into the molecules, in which the atom position $\mathbf{p}$ and the node features $\mathbf{h}$ are utilized for watermark embedding. The embedding process is defined as:
\begin{equation}
\mathbf{p}' = \mathcal{E}_\phi(\mathbf{p},\mathbf{h},\mathbf{m}),
\end{equation}
where $\mathbf{p}' = (\mathbf{p}'_1, ..., \mathbf{p}'_N) \in \mathbb{R}^{N \times 3}$ is the watermarked atom positions. Specifically, the encoder only modifies the atom positions while keeping the node features unchanged. In this way, the watermarked molecular geometry is expressed as $\mathcal{G}'=\left\langle {\mathbf{p}',\mathbf{h}} \right\rangle  \in \mathbb{R}^{N \times (3+d)}$.

\begin{figure*}[htbp]
  \centering
  \includegraphics[width=0.8\textwidth]{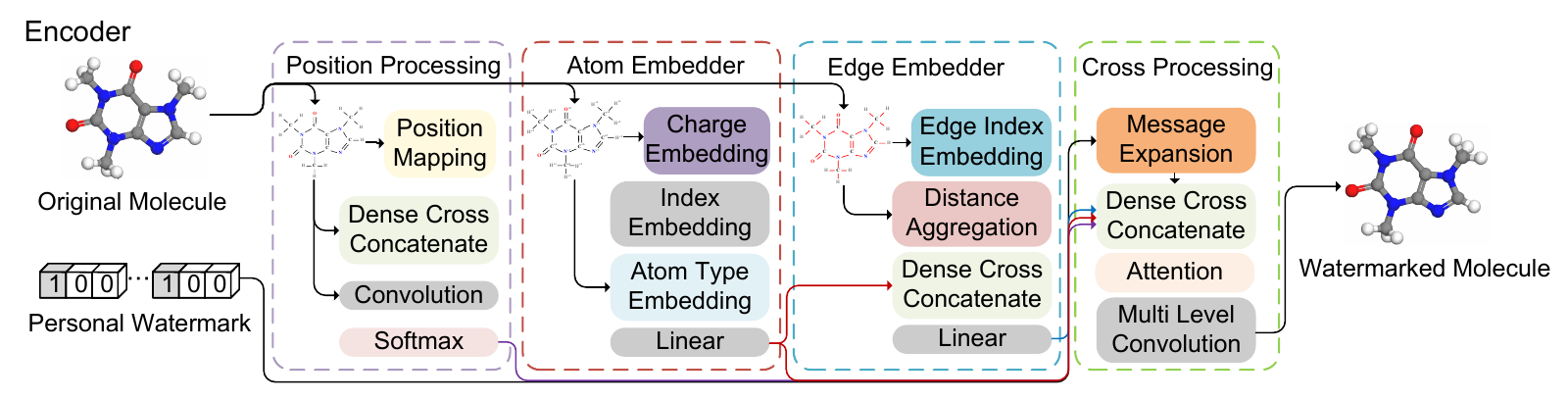}
  \caption{The detailed structures of encoder $\mathcal{E}_\phi$, including the position processing module, atom embedder, edge embedder, cross processing module. The atom embedder and the edge embedder effectively utilize the atom-level features, ensuring the effectiveness of watermarked molecules.}
  \label{Fig3}
\end{figure*}

\subsubsection{Structure of Encoder}
\label{section:Structure of Encoder}

Fig.~\ref{Fig3} presents the structure of the encoder $\mathcal{E}_\phi$, including the position processing module, the atom embedder, the edge embedder, and the cross processing module. The functionalities of these modules are summarized below.

\textbf{\emph{Position Processing Module:}} Since each molecule contains only a small number of atoms, the available redundancy for watermark embedding is limited. To alleviate this constraint, the position processing module projects the atom positions $\mathbf{p}$ into a latent space, producing the position features $\mathbf{f}_p$:
\begin{equation}
\mathbf{f}_p = f_{position}({\mathbf{p}}).
\end{equation}
Given the atom positions $\mathbf{p} = (\mathbf{p}_1, ..., \mathbf{p}_N) \in \mathbb{R}^{N \times 3}$, we first construct an atom-position batch $\mathbf{p}^B \in \mathbb{R}^{B \times 1 \times N \times 3}$, where the batch size is $B=64$. Several convolution operations are applied to transform $\mathbf{p}^B$ into position features $\mathbf{f}_p \in \mathbb{R}^{B \times C \times N \times e}$, with the channel dimension $C=64$. This module constructs spatial representation and introduces redundancy, enabling effective watermark embedding.

\textbf{\emph{Atom Embedder:}} The atom embedder leverages the node features $\mathbf{h}$ to facilitate watermark embedding. In particular, the atom type $\mathbf{t}$ and atom charge $\mathbf{c}$ are incorporated to generate the atom features $\mathbf{f}_a$:
\begin{equation}
\mathbf{f}_a = f_{atom}(\mathbf{t},\mathbf{c}).
\end{equation}
For a batch of molecules, the atom type batch $\mathbf{t}^B \in \mathbb{R}^{B \times 1 \times N \times e}$ and atom charge batch $\mathbf{c}^B \in \mathbb{R}^{B \times 1 \times N \times 1}$ are constructed. To enrich these representations, we adopt the sinusoidal positional encoding from Transformer~\cite{vaswani2017attention} to calculate the embedding $PE$ with length $N$:
\begin{equation}
PE_{(pos,i)} = \left\{ \begin{array}{l}
\sin (\frac{{pos}}{{{num^{i/{d_{model}}}}}}) \;\;\;\;\;\;\;\; \text{if} \;\; mod(i,2) =  = 0,\\
\cos (\frac{{pos}}{{{num^{(i - 1)/{d_{model}}}}}})\;\;\text{if} \;\; mod(i,2) =  = 1,
\end{array} \right.
\end{equation}
where $0 \le i < d_{model} / 2, d_{model}=64$, $pos$ is the index of the position, and $num$ is the maximum length of the position embedding. The resulting embedding is expanded to the batch size, yielding $PE \in \mathbb{R}^{B \times 1 \times N \times d_{model}}$. Following this, the atom charge bath $\mathbf{c}^B$ is mapped as follows:
\begin{equation}
\mathbf{c}^B = PE \circ \mathbf{c}^B,
\end{equation}
where $\circ$ calculate the Hadamard product to produce $\mathbf{c}^B \in \mathbb{R}^{B \times 1 \times N \times d_{model}}$. Meanwhile, the atom type batch $\mathbf{t}^B$ is linearly projected into the same latent dimension. The transformed atom type features and charge features are then concatenated along the last dimension, thereby generating the final atom features $\mathbf{f}_a \in \mathbb{R}^{B \times 1 \times N \times 3}$.

\textbf{\emph{Edge Embedder:}} To further enhance the watermark embedding process, the edge embedder explicitly utilizes interatomic correlations. Molecular structures are highly sensitive to spatial perturbations that small deviations in atomic coordinates may substantially violate chemical bond constraints. To this end, the edge embedder incorporates interatomic distances to produce the edge features $\mathbf{f}_e$:
\begin{equation}
\mathbf{f}_e = f_{edge}(\mathbf{e}, \mathbf{f}_a, \mathbf{p}).
\end{equation}
In this way, the edge index batch is represented as $\mathbf{e}^B \in \mathbb{R}^{B \times 1 \times N \times (N-1)}$. To enhance robustness against spatial perturbations, we construct an aggregated descriptor:
\begin{equation}
\mathbf{a}_d = f_{agg}(\mathbf{e},\mathbf{p}),
\end{equation}
where $f_{agg}$ is the aggregation methods such as the summation (``SUM'') and averaging (``MEAN''). The resulting $\mathbf{a}_d \in \mathbb{R}^{B \times 1 \times N \times 1}$ is concatenated with the atom features $\mathbf{f}_a$ along the last dimension, and the combined representations are further transformed through a linear layer to produce the aggregated atom features $\mathbf{f}'_a$.

Following this, the position embedding $PE$~\cite{vaswani2017attention} is concatenated with the aggregated features $\mathbf{f}'_a$. This concatenated representation is passed through several linear layers with layer normalization, yielding the final edge feature $\mathbf{f}_e \in \mathbb{R}^{B \times 1 \times N \times 3}$.

\textbf{\emph{Cross Processing Module:}} Based on the features, the cross processing module integrates the position features $\mathbf{f}_p$, the atom features $\mathbf{f}_a$, and the edge features $\mathbf{f}_e$ with the watermark $\mathbf{m}$, generating the watermarked atom positions $\mathbf{p}' = (\mathbf{p}'_1, ..., \mathbf{p}'_N) \in \mathbb{R}^{N \times 3}$:
\begin{equation}
\mathbf{p}' = f_{cross}(\mathbf{f}_p, \mathbf{f}_a, \mathbf{f}_e, \mathbf{m}).
\end{equation}
The watermark $\mathbf{m}$ is a binary sequence of length $L$, which is expanded along the spatial dimensions to form $\mathbf{m}^e \in \mathbb{R}^{B \times L \times N \times 3}$. The latent representation $\mathbf{f}_c$ is obtained by concatenating $\mathbf{f}_p$, $\mathbf{f}_a$, $\mathbf{f}_e$, and $\mathbf{m}^e$ along the channel dimension. This composite tensor is processed through three cross operations~\cite{huang2017densely} and a self-attention mechanism~\cite{vaswani2017attention} to generate the feature mask $\mathbf{f}_m$.

Following this, the feature mask $\mathbf{f}_c$ is element-wise multiplied with $\mathbf{f}_m$, and the result is passed by a convolution layer to produce a position mask $\mathbf{p}_m \in \mathbb{R}^{B \times 1 \times N \times 3}$. Finally, the watermarked atom positions $\mathbf{p}'$ are obtained as:
\begin{equation}
\mathbf{p}' = \mathbf{p} + \mathbf{p}_m.
\end{equation}
In this way, these strategies enhance feature interaction and ensure that the watermark is deeply embedded in an invasive way.

\subsubsection{Optimization of the Encoder}
\label{section:Optimization of the Encoder}

The four components of the encoder $\mathcal{E}_\phi$ collaboratively embed the watermark $\mathbf{m}$ into the atom position $\mathbf{p}$, in which the node features $\mathbf{h}$ are utilized to generate the watermarked molecules. To reduce perturbation to molecular structures and preserve chemical properties, the embedding process is guided by a loss function that minimize deviations between the original and watermarked molecules. Specifically, the encoder loss $\mathcal{L}_E$ is defined as:
\begin{equation}
{\mathcal{L}_E}(\mathbf{p}, \mathbf{p}') = \max (\left| {\mathbf{p} - \mathbf{p}'} \right|^2) + \frac{1}{N}\sum\limits_{i = 1}^N {\left\| {\mathbf{p}_i - \mathbf{p}'_i} \right\|_2^2},
\label{Eq9}
\end{equation}
where $\max(\cdot)$ denotes the maximum element of the input matrix and $\left\|{\cdot}\right\|_2^2$ represents the squared Euclidean norm. The encoder loss $\mathcal{L}_E$ is composed of two complementary components:
\begin{enumerate}
  \item \textbf{Maximum Distance:} Captures the largest positional shift among all atoms, providing a strict upper bound of the distortion. However, this loss is non-differentiable that cannot be directly optimized using gradient-based methods.
  \item \textbf{Euclidean Norm:} Measures the overall displacement using the Euclidean norm $\ell_2$, which is differentiable and thus enables effective gradient descent during training.
\end{enumerate}

By combining these two complementary components, the loss function ensures that the encoder parameters $\phi$ are optimized to minimize structural perturbations. This design ensures that the encoder can effectively embed watermarks while minimally altering molecular structures, thereby preserving molecular properties.

\begin{figure*}[!h]
  \centering
  \includegraphics[width=0.8\textwidth]{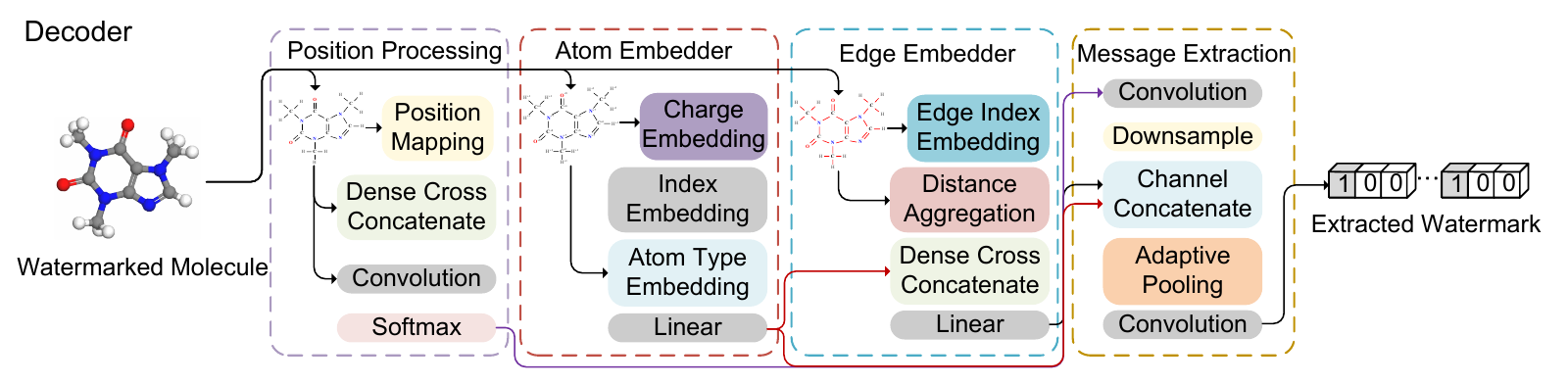}
  \caption{The detailed structures of decoder $\mathcal{D}_\theta$, including the position processing module, atom embedder, edge embedder, and message extraction module. The atom embedder and edge embedder share the same structures as the counterparts in the encoder, but are trained with independent parameters.}
  \label{Fig4}
\end{figure*}

\subsection{Watermark Extraction}
\label{section:Watermark Extraction}

The decoder $\mathcal{D}_\theta$ is employed to extract the embedded watermark from the watermarked molecules $\mathcal{G}'=\left\langle {\mathbf{p}',\mathbf{h}} \right\rangle  \in \mathbb{R}^{N \times (3+d)}$. Since molecular properties remain unchanged under SE(3) transformations, the processed molecules should be protected. This requirement implies that \OURS{} must be robust against SE(3) transformations, indicating the decoder $\mathcal{D}_\theta$ can extract watermark when molecules are manipulated or visualized in arbitrary spatial configurations.

To guarantee robustness, we figure out the SE(3)-invariant features from molecules $\mathcal{G}=\left\langle {\mathbf{p},\mathbf{h}} \right\rangle  \in \mathbb{R}^{N \times (3+d)}$. The node features $\mathbf{h}$ are scalar that invariant under SE(3) transformations while the atom positions $\mathbf{p}$ are not. Thus, the invariant features are constructed from atom positions $\mathbf{p}$, facilitating the decoder $\mathcal{D}_\theta$ to extract the embedded watermark. The SE(3)-invariant features must satisfy the following criteria:
\begin{equation}
\mathcal{F}(\mathbf{p}) = \mathcal{F}(\mathbf{p}\mathbf{A} + \mathbf{T}),
\label{Eq10}
\end{equation}
where $\mathcal{F}(\cdot)$ denotes the feature extraction function, $\mathbf{A}$ is a linear transformation matrix (e.g., rotation or reflection), and $\mathbf{T}$ is a translation vector. In our experiments, we figure out that Euclidean distances between atom positions satisfy the invariance property. The resulting distance matrix captures the internal geometry of the molecules while discarding the absolute orientations and global positions. Accordingly, the Euclidean distance matrix $\mathbf{d} \in \mathbb{R}^{N \times N}$ is calculated as the SE(3)-invariant features. For the $i$-th and $j$-th atoms, the distance $d_{i,j}$ is calculated as follows:
\begin{equation}
{d_{i,j}} = \sqrt {{{({x_i} - {x_j})}^2} + {{({y_i} - {y_j})}^2} + {{({z_i} - {z_j})}^2}}
\label{Eq11}
\end{equation}
where $(x_i, y_i, z_i)$ and $(x_j, y_j, z_j)$ are the coordinates of the $i$-th and the $j$-th atoms, respectively.

Although $\mathbf{d}$ provides a rich and stable invariant representation, directly extracting watermark can introduce numerical instability, which may impair decoder performance. To mitigate this issue, we employ the multidimensional scaling (MDS)~\cite{carroll1998multidimensional} to reconstruct coordinate features $\hat{\mathbf{p}} \in \mathbb{R}^{N \times 3}$ that approximate the watermarked positions $\mathbf{p}'$. This reconstruction preserves SE(3) invariance while enabling the decoder to operate in a geometry-aware coordinate space, thereby improving extraction accuracy and stability.

\subsubsection{Multi Dimensional Scaling}
\label{section:Multi Dimensional Scaling}

The MDS method are described as follows. We first construct the centered distance matrix $\mathbf{B} \in \mathbb{R}^{N \times N}$:
\begin{equation}
\mathbf{B} = -\frac{1}{2} \mathbf{J} \mathbf{d}^2 \mathbf{J},
\end{equation}
where $\mathbf{d}^2$ denotes element-wise squared distances. The centered matrix $\mathbf{J} \in \mathbb{R}^{N \times N}$ is defined as follows:
\begin{equation}
\mathbf{J} = \mathbf{I} - \frac{1}{n} \mathbf{1} \mathbf{1}^T.
\end{equation}
Here, $\mathbf{I} \in \mathbb{R}^{N \times N}$ represents the identity matrix and $\mathbf{1} \in \mathbb{R}^{N \times 1}$ denotes the matrix of ones. This double centering step removes global translation components, yielding an inner product equivalent representation suitable for coordinate reconstruction. After that, the eigenvalue decomposition is applied to the centered matrix:
\begin{equation}
\mathbf{B} = \mathbf{V} {\Lambda} \mathbf{V}^T,
\end{equation}
where ${\Lambda} \in \mathbb{R}^{N \times N}$ is the eigenvalue matrix and $\mathbf{V} \in \mathbb{R}^{N \times N}$ is the eigenvector matrix.

In this way, the top $k$ positive eigenvalues and the corresponding eigenvectors are used to recover the atom position $\mathbf{\hat p}$ as follows:
\begin{equation}
\mathbf{\hat p} = \mathbf{V}_k {\Lambda}_k^{1/2},
\end{equation}
where ${\Lambda}_k \in \mathbb{R}^{k \times k}$ is the diagonal matrix of the top $k$ eigenvalues, and $\mathbf{V}_k \in \mathbb{R}^{k \times k}$ is the matrix containing the corresponding eigenvectors. We set $k=3$ to ensure that $\hat{\mathbf{p}}$ aligns with the watermarked positions $\mathbf{p}'$.

\subsubsection{Structure of Decoder}
\label{section:Structure of Decoder}

After converting the distance matrix $\mathbf{d}$ into reconstructed positions $\mathbf{\hat p}$ via the MDS method, the decoder $\mathcal{D}_\theta$ is employed to extract the embedded watermark $\mathbf{m}' = (\mathbf{m}'_1, ..., \mathbf{m}'_L) \in \mathbb{R}^{1 \times L}$. The extraction process is expressed as follows:
\begin{equation}
\mathbf{m}' = \mathcal{D}_\theta(\mathbf{\hat p}, \mathbf{h}),
\end{equation}
where the node features $\mathbf{h}$ provide complementary information to facilitate reliable decoding.

As shown in Fig.~\ref{Fig4}, the decoder $\mathcal{D}_\theta$ consists of four components, including position mapping module, atom embedder, edge embedder, and extraction module. The atom embedder and edge embedder adopt the same structures as the counterparts in encoder, but are trained with independent parameters. Therefore, more focus is applied on describing the extraction module, which recovers the embedded watermark from the processed molecules.

\textbf{\emph{Extraction Module:}} Aiming to extract the embedded watermark with high bit accuracy. The preceding components provide the embedded position features $\mathbf{f}'_p$, atom features $\mathbf{f}_a$, and edge features $\mathbf{f}_e$. The watermark $\mathbf{m}'$ is extracted as follows:
\begin{equation}
\mathbf{m}' = f_{extract}(\mathbf{f}'_p, \mathbf{f}_a, \mathbf{f}_e).
\end{equation}

Specifically, the three features ($\mathbf{f}'_p$, $\mathbf{f}_a$, and $\mathbf{f}_e$) are first concatenated along the channel dimension to form a unified latent representation $\mathbf{f}_c$. This representation is then processed by three cross operations, enabling reliable watermark extraction. To match the target watermark length $L$, an adaptive average pooling strategy reshapes the output features into a tensor of size $(B, L, 1, 1)$. Finally, an element-wise rounding operation is applied to the pooled outputs, yielding the binarized watermark $\mathbf{m}'$.

\subsubsection{Optimization of the Decoder}
\label{section:Optimization of the Decoder}

Based on the parameterized decoder $\mathcal{D}_\theta$, the watermark $\mathbf{m}'$ can be extracted. To improve the decoding performance, the discrepancy between the extracted watermark $\mathbf{m}'$ and the original watermark $\mathbf{m}$ is measured by the Euclidean norm $\ell_2$, yielding the decoder loss $\mathcal{L}_D$:
\begin{equation}
\mathcal{L}_D(\mathbf{m},\mathbf{m}') = \frac{1}{L}\sum\limits_{i = 1}^L {\left\| {\mathbf{m}_i - \mathbf{m}'_i} \right\|_2^2}.
\end{equation}

During training, $\mathcal{L}_D$ is minimized to optimize the decoder parameters $\theta$, improving the accuracy of watermark extraction through a learnable strategy.

\subsection{Training Objective}
\label{section:Training Objective}

The objectives of \OURS{} are maintaining the molecular properties after embedding watermarks and improving the bit accuracy of the extracted watermarks. We perform Adam optimization~\cite{kingma2014adam} with default hyperparameters to minimize the weighted sum of the encoder and decoder losses over the distribution of molecules and watermarks:
\begin{equation}
\mathop {\min }\limits_{\phi ,\theta } {\mathbb{E}_{(\mathbf{p},\mathbf{m})}}\left[ {{\lambda _E}{\mathcal{L}_E}(\mathbf{p},\mathbf{p}') + {\lambda _D}{\mathcal{L}_D}(\mathbf{m},\mathbf{m}')} \right],
\label{Eq19}
\end{equation}
where $\lambda _E$ and $\lambda _D$ are the relative weights of the encoder loss and the decoder loss, respectively.

To achieve the training objectives, we introduce a dynamic weighting strategy. Specifically, $\lambda _E$ increases gradually with training progress:
\begin{equation}
{\lambda _E} = \rho  + \beta  \times \left\lfloor {\frac{{t}}{{f}}} \right\rfloor,
\label{Eq20}
\end{equation}
where $t$ is the current training epoch. Additionally, the initial weight is $\rho=0.01$, the increase factor is $\beta=0.25$, and the increase interval is $f=50$. Meanwhile, the decoder weight is modulated by the real-time bit accuracy $\gamma$:
\begin{equation}
{\lambda _D} = \delta  \times (1 - \gamma),
\label{Eq21}
\end{equation}
where the increase factor is $\delta=100$. At the beginning of training, $\lambda_D$ is significantly larger than $\lambda_E$, guiding \OURS{} to focus on improving the decoder for watermark extraction. As $\gamma$ increases, $\lambda_D$ decreases accordingly, shifting \OURS{} to refine the encoder and preserving molecular structures. Finally, \OURS{} can achieve robust watermark embedding while preserving molecular properties.


\begin{table*}[htbp]
\caption{Feasibility of embedding different capacities of watermarks on molecules by using \OURS{}. The molecules originate from QM9 dataset, GEOM-DRUG dataset, and GeoBFN model, which are embedded watermarks using \OURS{} to evaluate the basic properties.}
\centering
\setlength{\tabcolsep}{1.2mm}
\begin{tabular}{ll|cccccc|ccc}
\toprule
\multicolumn{2}{c|}{\multirow{2}{*}{Methods}} & \multicolumn{6}{c|}{QM9}                                   & \multicolumn{3}{c}{GEOM-DRUG} \\
\multicolumn{2}{c|}{}                         & Atom Sta & Mol Sta & Validity & Uniq  & Novelty & Bit Acc & Atom Sta & Validity & Bit Acc \\ \midrule
\multicolumn{2}{c|}{Data}                     & 99.00    & 95.20   & 97.70    & 97.70 & -       & -       & 86.50    & 99.90    & -       \\
\multicolumn{2}{c|}{GeoBFN}                   & 99.72    & 95.00   & 97.65    & 85.22 & 72.99   & -       & 86.25    & 92.50    & -       \\ \midrule
\multirow{5}{*}{MolMark}      & 1 bit        & 98.41    & 95.00   & 97.54    & 85.20 & 72.95   & 99.69   & 86.12    & 92.08    & 99.82   \\
                              & 4 bits       & 98.12    & 95.00   & 97.50    & 85.23 & 73.02   & 97.66   & 85.89    & 91.28    & 97.49   \\
                              & 8 bits       & 98.07    & 94.69   & 97.45    & 85.20 & 73.08   & 95.59   & 84.96    & 90.03    & 95.61   \\
                              & 12 bits      & 97.96    & 94.62   & 97.33    & 85.91 & 73.49   & 95.14   & 83.78    & 88.06    & 95.75   \\
                              & 16 bits      & 97.69    & 94.61   & 97.36    & 85.22 & 72.99   & 95.20   & 83.49    & 86.06    & 95.05   \\
\bottomrule
\label{Table2}
\end{tabular}
\end{table*}

\begin{figure*}[htbp]
  \centering
  \includegraphics[width=1\textwidth]{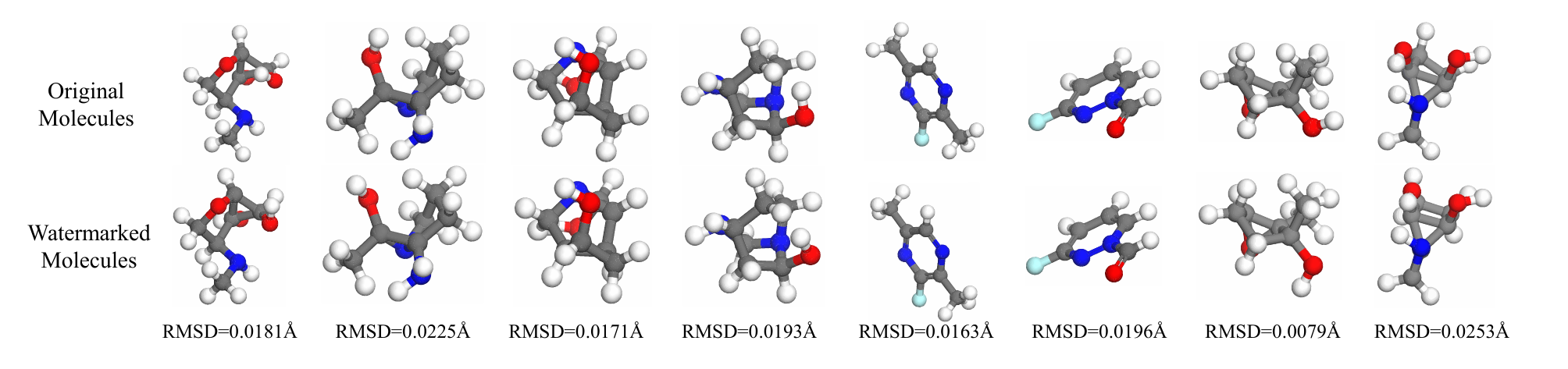}
  \caption{Structure of eight pairs of molecules. The original molecule and watermarked molecules are arranged vertically, in which the structures only undergo slight changes after embedding watermarks. The structural differences are minimal, in which all RMSD values lower than 0.03~\AA.}
  \label{Fig5}
\end{figure*}

\section{Experimental Results}
\label{section:Experimental Results}

In this section, we conduct several experiments to assess the performance of \OURS{} by using molecules from the two widely adopted benchmark datasets (QM9~\cite{ramakrishnan2014quantum} and GEOM-DRUG~\cite{axelrod2022geom}), and two state-of-the-art unconditional 3D molecular generative models (GeoBFN~\cite{song2023unified} and GeoLDM~\cite{xu2023geometric}). QM9 dataset contains approximately 134,000 small organic molecules while GEOM-DRUG containing roughly 450,000 drug-like molecules and 37 million geometry-relaxed conformations. Following the established protocols in prior works~\cite{anderson2019cormorant,hoogeboom2022equivariant,xu2023geometric}, we use a standardized split of 100,000 molecules for training, 18,000 for validation, and 13,000 for testing.

The performance of \OURS{} is evaluated through a series of experiments. We first explore the feasibility of embedding watermarks into molecules, establishing a baseline for evaluating embedding performance. We then analyze the influence of watermark capacities on molecular properties. Additionally, comparisons with existing robust watermarking methods emphasize the necessity of designing \OURS{} for molecular protection. The functionality of watermarked molecules is further assessed by analyzing physicochemical properties and downstream docking performance. Robustness to geometric perturbations is evaluated under the SE(3) transformations, including rotations, translations, and reflections. Finally, ablation studies quantify the contributions of key components within \OURS{}, validating the effectiveness of the method design.

\subsection{Feasibility of Embedding Watermarks in Molecular Structures}
\label{Section:Feasibility of embedding watermarks into molecules}

We first investigate the feasibility of embedding watermarks into molecular structures. The watermark $\mathbf{m}$ is defined as a binary sequence with length $L$, allowing up to $2^L$ distinct identifiers. Although watermarking methods have been successfully applied to protect images and proteins, directly transferring these techniques to molecules introduces a fundamental mismatch. As shown in Table~\ref{Table1}, previous watermarking methods~\cite{zhu2018hidden,liu2019novel,huang2023arwgan} embed about 30 bits into images with $128 \times 128 \times 3$ pixels, yielding a bit-per-pixel (BPP) of $1.83 \times 10^{-3}$. Similarly, protein watermarking methods embed 32 bits into structures containing approximately $500 \times 37$ atoms~\cite{zhang2024foldmark}, yielding a bit-per-atom (BPA) of $1.73 \times 10^{-3}$. In contrast, molecules in datasets such as QM9 and GEOM-DRUG generally contain no more than 200 atoms. Given the compactness of molecular structures and their strict bonding constraints, designing a new watermarking method for molecules is inherently challenging.

Maintaining the similar BPA used in protein watermarking~\cite{zhang2024foldmark} yields a theoretical watermark length of $L = 0.35$ bits. Since the watermark length must be an integer, we set $L = 1$ to evaluate the feasibility of \OURS{} in embedding watermark into molecules. To quantify feasibility, we utilize the basic properties of molecules and the bit accuracy of the embedded watermark as evaluation metrics. Following prior works~\cite{hoogeboom2022equivariant,wu2022diffusion,xu2023geometric,song2023unified}, basic molecular properties include atom stability (Atom Sta), molecule stability (Mol Sta), validity, uniqueness (Uniq), and novelty. Atom and molecule stabilities evaluate bonding correctness and global structural coherence. Validity captures adherence to chemical rules, while uniqueness and novelty assess redundancy and dataset overlap. The bit accuracy (Bit Acc) is defined as the percentage of correctly recovered bits. Overall, higher values across these metrics reflect better feasibility.

Table~\ref{Table2} reports the results of embedding 1-bit watermark into molecules from QM9 and GEOM-DRUG datasets. For QM9 dataset, the basic properties change slightly, in which atom stability, molecule stability, and validity decrease from 99.00\%, 95.20\%, and 97.70\% to 98.41\%, 95.00\%, and 97.54\%, respectively. The bit accuracy reaches 99.69\%, demonstrating highly reliable watermark extraction. For GEOM-DRUG dataset, a mild degradation in basic properties is observed, where atom stability drops from 86.50\% to 86.12\% and validity from 99.90\% to 92.08\%. Also, the bit accuracy remains high at 99.82\%, confirming that \OURS{} is feasible for embedding watermark into molecules.

Fig.~\ref{Fig5} visualizes eight representative original and watermarked molecules. The structural differences are minimal, with root-mean-square deviations (RMSD) below 0.03~\AA. Only slight perturbations occur on hydrogen atoms, while the chemical bonds remain essentially unchanged. Based on the high molecular properties and bit accuracies, we can demonstrate that \OURS{} can effectively embed watermarks into molecules.

\subsection{Investigation of Watermark Embedding Capacity in Molecular Properties}

Section~\ref{Section:Feasibility of embedding watermarks into molecules} has demonstrated the feasibility of \OURS{} in embedding 1-bit watermark into molecules without compromising molecular properties. However, the practical utility is limited by the low capacity. Increasing the watermark length is essential, but larger capacities generally introduce stronger perturbations. To quantify this trade-off, we set the embedding capacity $L$ from 4 to 16 bits, achieving a balance between copyright protection and structural preservation.

Table~\ref{Table2} presents the impact of \OURS{} on embedding different watermarks on molecules generated by GeoBFN. Compared to the original molecules, the watermarked molecules exhibit minor decline in basic properties. Atom stability decreases slightly from 99.72\% but remain above 98\% across all capacities, and molecule stability shows a modest reduction. Validity, uniqueness, and novelty remain higher than 90\%, indicating that \OURS{} can maintain structural coherence as more watermarks are embedded. Novelty rises gradually, suggesting that the subtle perturbations introduced by \OURS{} promote greater conformational diversity. Bit accuracies remain above 95\%, confirming that the watermarks can be extracted accurately at higher capacities. These results underscore the minimal impact of \OURS{} on molecules across a range of capacities. Additional evaluations on GeoLDM~\cite{xu2023geometric} further confirm these findings, as reported in Supplementary Section~1.1.

Collectively, the experiments across multiple watermark capacities show that \OURS{} can protect molecules while preserving molecular properties. This balance between extraction accuracy of watermark and structure integrity of molecules supports both copyright enforcement and downstream traceability. As generative models become increasingly influential in molecular discovery, \OURS{} offer a practical path toward secure, transparent, and responsible molecular innovation.

\begin{table*}[thbp]
\caption{Feasibility of embedding different capacities of watermarks on molecules by using \OURS{}. The molecules originate from QM9 dataset, GEOM-DRUG dataset, and GeoBFN model, in which \OURS{} embeds watermarks for evaluating the basic properties.}
\centering
\setlength{\tabcolsep}{1.2mm}
\begin{tabular}{lcccccc|ccc}
\toprule
\multicolumn{1}{c|}{\multirow{2}{*}{Methods}}  & \multicolumn{6}{c|}{QM9}                                   & \multicolumn{3}{c}{GEOM-DRUG} \\
\multicolumn{1}{c|}{}                          & Atom Sta & Mol Sta & Validity & Uniq  & Novelty & Bit Acc & Atom Sta & Validity & Bit Acc \\ \midrule
\multicolumn{1}{c|}{MolMark}                  & 98.07    & 94.69   & 97.45    & 85.20 & 73.08   & 95.59   & 84.96    & 90.03    & 95.61   \\
\multicolumn{1}{c|}{HiDDeN}                   & 0.00     & 40.96   & 16.22    & 71.10 & 73.02   & 94.64   & 0.00     & 21.15    & 90.28   \\
\multicolumn{1}{c|}{MBRS}                     & 0.00     & 43.76   & 21.18    & 73.21 & 73.49   & 95.35   & 0.00     & 25.36    & 89.56   \\
\multicolumn{1}{c|}{ARWGAN}                   & 0.00     & 49.61   & 28.32    & 74.15 & 72.99   & 95.46   & 0.00     & 30.31    & 93.54  \\
\bottomrule
\label{Table3}
\end{tabular}
\end{table*}

\subsection{Comparison with Previous Watermarking Methods}

Watermarking methods have been extensively explored in several multimedia domains. HiDDeN~\cite{zhu2018hidden}, MBRS~\cite{jia2021mbrs}, and ARWGAN~\cite{huang2023arwgan} are three representative methods for traditional multimedia contents. Recently, watermarking methods have been applied to substances such as proteins~\cite{zhang2024foldmark,chen2025enhancing} and DNA~\cite{zhang2025securing,zhang2025genebreaker}, which exploit the stability of rigid structural elements. However, small molecules exhibit complex chemical bond topologies and diverse three-dimensional conformations, lacking directly manipulable linear sequence. Therefore, approaches designed for proteins or DNA are not applicable to small molecules.

Applying previous watermarking methods to molecules reveals further limitations. To investigate these limitations, we utilize HiDDeN~\cite{zhu2018hidden}, MBRS~\cite{jia2021mbrs}, and ARWGAN~\cite{huang2023arwgan} to embed watermarks into molecules, in which the models are trained on QM9 and GEOM-DRUG datasets and evaluated on molecules generated by GeoBFN, with embedding capacity of 16 bits. Table~\ref{Table3} compares \OURS{} with HiDDeN, MBRS, and ARWGAN in terms of molecular properties and bit accuracy. On the QM9 dataset, all methods achieve comparable bit accuracies around 95\%. However, HiDDeN, MBRS, and ARWGAN degrade molecular properties that molecule stability drops to 0\% and atom stability declines to 40-50\%, rendering molecules functionally ineffective. In contrast, \OURS{} consistently preserves molecular properties above 90\%, demonstrating its ability to embed watermarks without compromising structural integrity.

These results underscore the structural damage caused by previous watermarking methods in embedding watermark into molecules. In contrast, \OURS{} achieves a trade-off between providing copyright protection and maintaining molecular properties. As generative models increasingly drive molecular design, \OURS{} provides a robust foundation for secure and responsible molecular innovation.

\begin{figure}[t]
  \centering
  \includegraphics[width=0.45\textwidth]{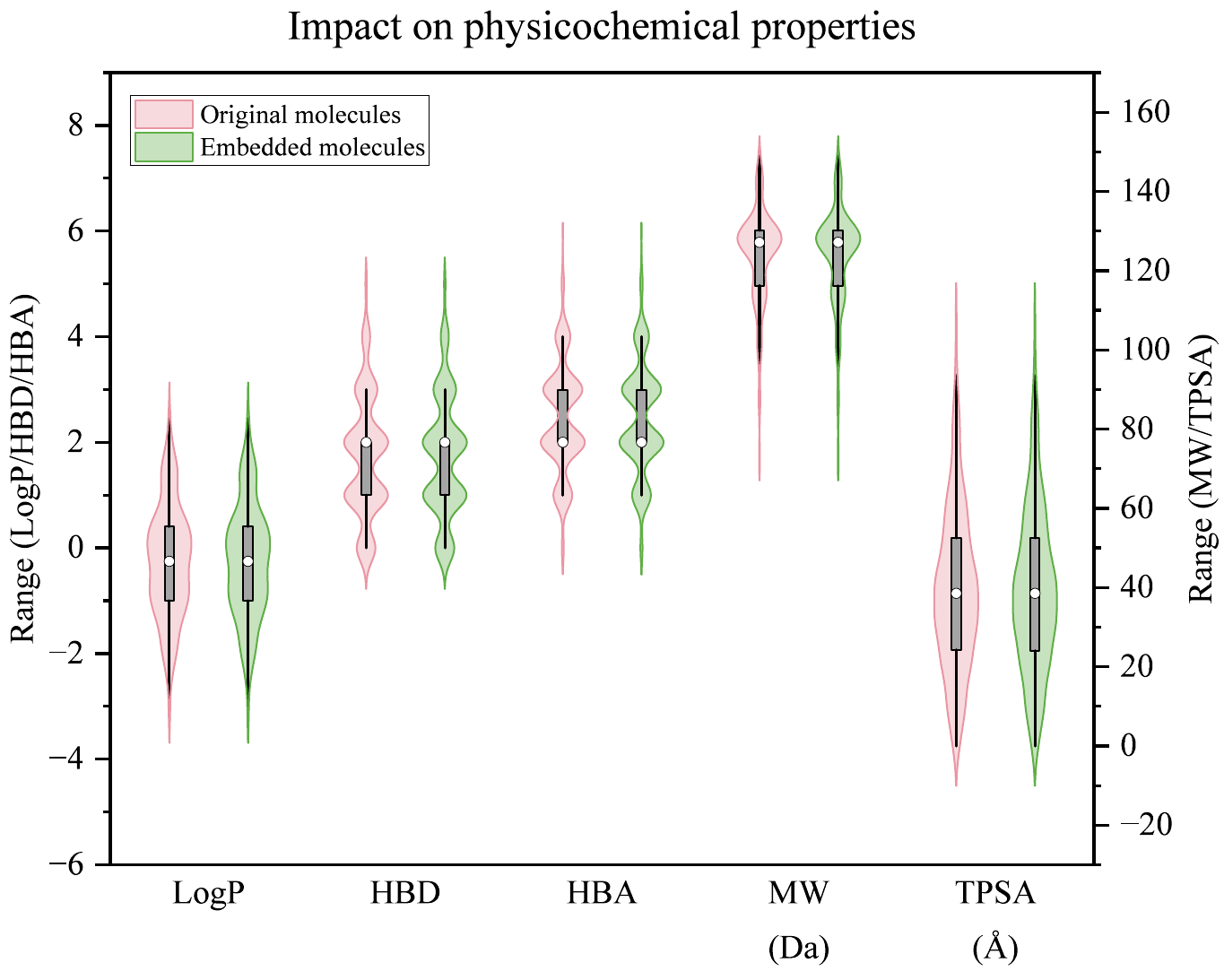}
  \caption{Impact of embedding watermark on molecular physicochemical properties. The original molecule and watermarked molecules are arranged vertically, in which the distributions of physicochemical properties are identical.}
  \label{Fig6}
\end{figure}

\subsection{Physicochemical Property Analysis of Watermarked Molecules}

In addition to evaluating basic properties, we further analyze the influences of embedding watermarks on the physicochemical properties of molecules. Structural rationality is first assessed using RDKit~\cite{landrum2006rdkit} and OpenBabel~\cite{o2011open}, which verify compliance with fundamental chemical rules, including bond valence, stereochemistry, and ring strain. Molecules that satisfy these structural checks are analyzed using RDKit and MDAnalysis~\cite{michaud2011mdanalysis,gowers2019mdanalysis}. The computed physicochemical properties include the lipid-water partition coefficient (LogP), hydrogen bond donors (HBD), hydrogen bond acceptors (HBA), molecular weight (MW), and topological polar surface area (TPSA). LogP measures the lipophilicity, with positive values indicating lipid solubility and negative values indicating hydrophilicity. HBD and HBA quantify hydrogen-bond-forming potential. MW is the sum of atom weights, given in Daltons (Da). Using Ertl's algorithm~\cite{ertl2000fast}, TPSA reflects the total surface area contributed by polar atoms, measured in \AA$^2$. These metrics provide a comprehensive assessment of \OURS{} on affecting the physicochemical properties of molecules.

We perform analyses on molecules generated by GeoBFN, after which the generated molecules are embedded with 8-bit watermarks using \OURS{}. Structural evaluation shows that 97.50\% of the original molecules and 95.30\% of the watermarked molecules remain chemically valid, with the invalid cases being due to carbon atoms exceeding the valence limit. For the structurally valid molecules, Fig.~\ref{Fig6} presents the distribution of physicochemical properties. These distributions are summarized using median, interquartile range (IQR), and whisker length. The median serves as a robust central statistic, the IQR captures the variability of the middle 50\% of samples, and the whiskers reflect the spread of non-outlier data. In general, smaller IQR and whisker values indicate more compact distributions. For readability, results are shown in the format ``median (IQR, whisker)''.

As illustrated in Fig.~\ref{Fig6}, the physicochemical properties distributions of the original and watermarked molecules are identical. Both the original and watermark molecules exhibit identical LogP values of -0.25 (1.40, 5.50), indicating that \OURS{} can preserve the hydrophilic-lipophilic balance. The HBD and HBA values remain consistent at 2 (1, 3), confirming that functional group distributions are identical. The MW distribution centers around 127.14 (14.02, 53.08) Da, and the TPSA values remain at 38.55 (28.22, 90.52)~\AA$^2$, suggesting the watermarked molecules retained membrane-permeability characteristics. These aligned distributions demonstrate that \OURS{} maintains physicochemical integrity and does not interfere with downstream analyses. More investigation across different embedding capacities is provided in Supplementary Section~1.2.

\begin{figure}[t]
  \centering
  \includegraphics[width=0.45\textwidth]{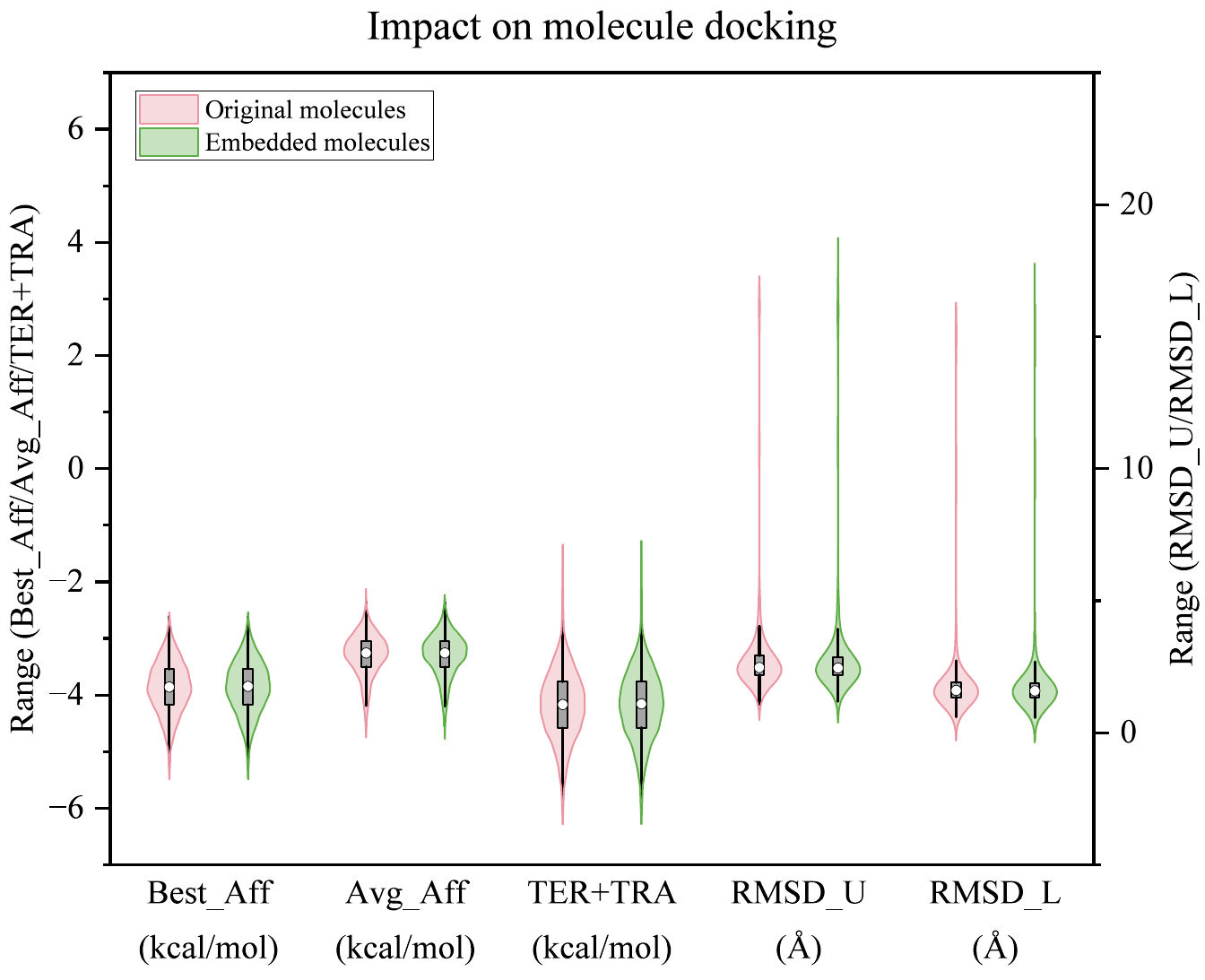}
  \caption{Impact of embedding watermark on molecule docking. The distributions of evaluation metrics of original molecules and the watermarked molecules are similar, showing that \OURS{} can maintain the functionality of molecules in downstream applications.}
  \label{Fig7}
\end{figure}

\begin{figure*}[htbp]
\centering
\subfigure[]
{
	\begin{minipage}[t]{0.48\linewidth}
    \centering
	\includegraphics[width=1\textwidth]{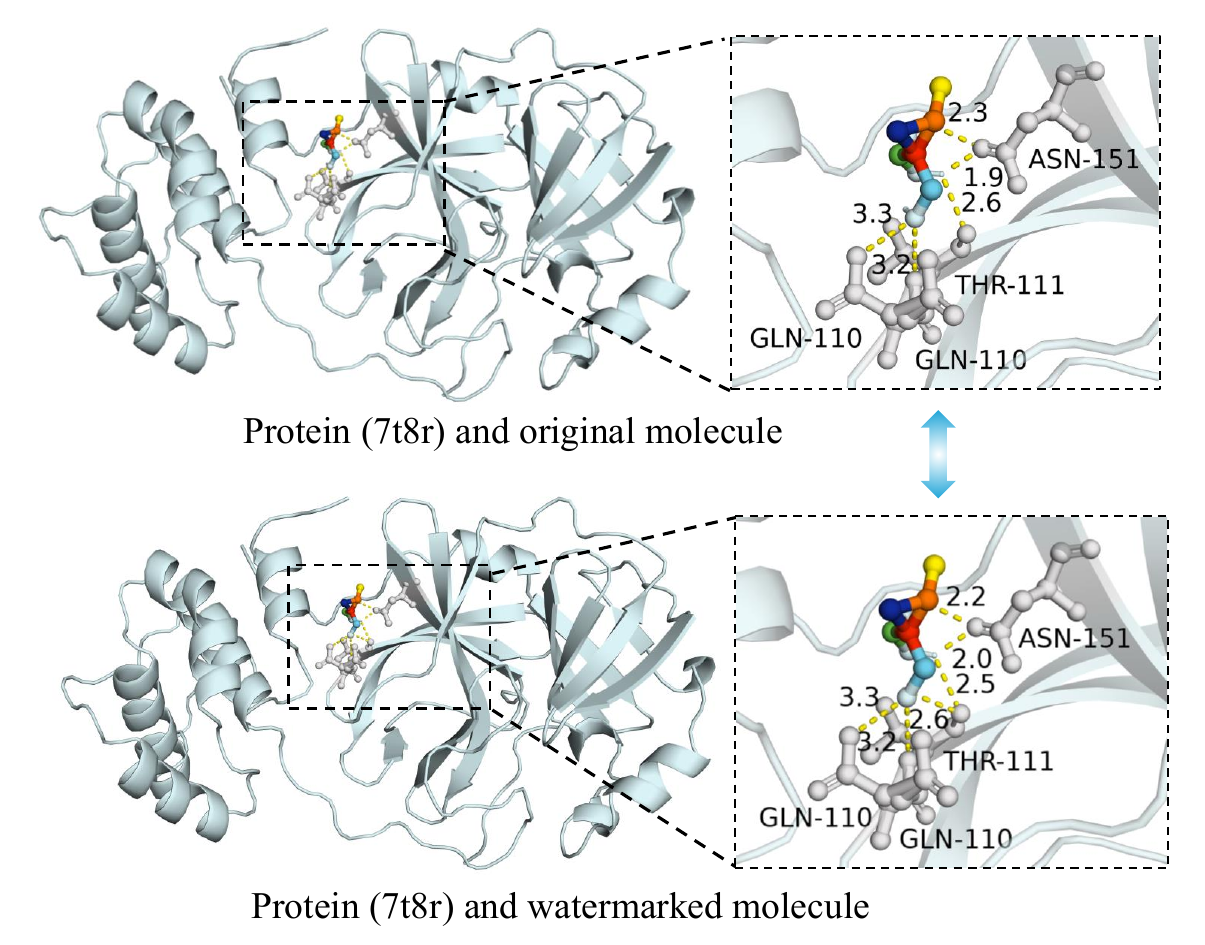}
	\end{minipage}
}
\subfigure[]
{
	\begin{minipage}[t]{0.48\linewidth}
    \centering
	\includegraphics[width=1\textwidth]{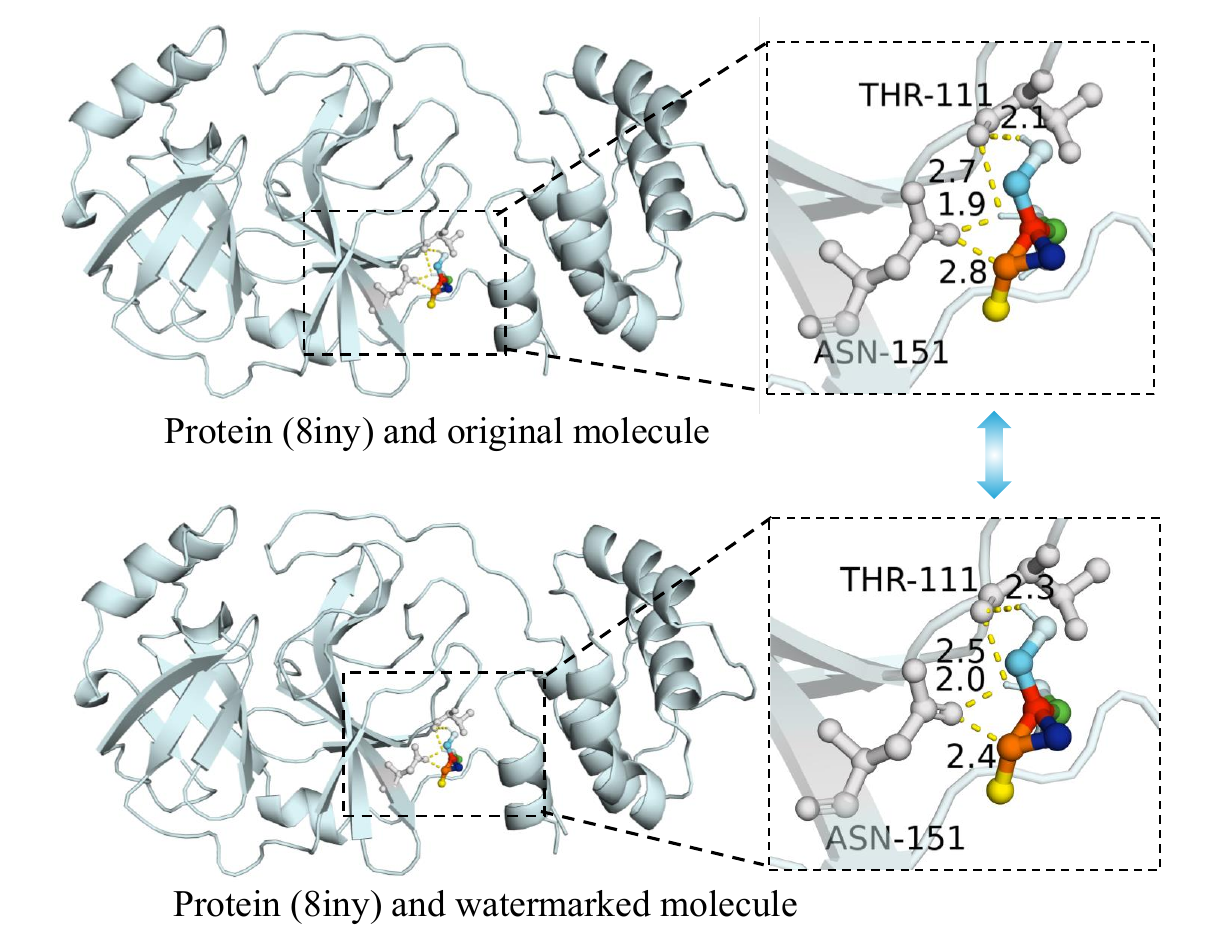}
	\end{minipage}
}
\caption{The docking rendering of original molecules and watermarked molecules with different proteins: (a) protein with PDB ID 7t8r, and (b) protein with PDB ID 8iny. The amino acid residues and hydrogen bond lengths in molecular docking are similar, reflecting that \OURS{} has an insignificant impact on the functionalities of molecules.}
\label{Fig8}
\end{figure*}

\begin{figure}[htbp]
\centering
\subfigure[]
{
	\begin{minipage}[t]{0.22\linewidth}
    \centering
	\includegraphics[width=0.9\textwidth]{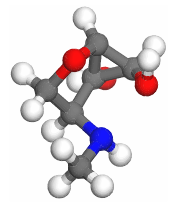}
	\end{minipage}
}
\subfigure[]
{
	\begin{minipage}[t]{0.22\linewidth}
    \centering
	\includegraphics[width=0.9\textwidth]{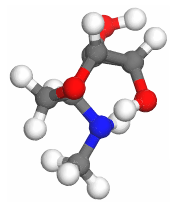}
	\end{minipage}
}
\subfigure[]
{
	\begin{minipage}[t]{0.22\linewidth}
    \centering
	\includegraphics[width=0.9\textwidth]{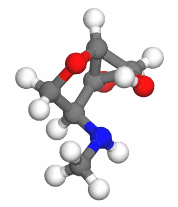}
	\end{minipage}
}
\subfigure[]
{
	\begin{minipage}[t]{0.22\linewidth}
    \centering
	\includegraphics[width=1\textwidth]{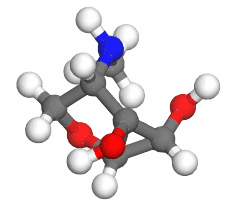}
	\end{minipage}
}
\caption{The conformations of molecule after applying SE(3) transformations: (a) original molecule, (b) rotated molecule, (c) translated molecule, and (d) reflected molecule. The conformations of the processed molecules are different, but they share the same characteristics.}
\label{Fig9}
\end{figure}

\subsection{Impact on Downstream Molecular Docking}

To evaluate whether \OURS{} affects downstream scientific analysis, we perform molecule docking as a representative task. We generate 100 molecules using GeoBFN and embed 8-bit watermarks into each molecule using \OURS{}, generating 200 molecules as the docking ligands. For the docking receptors, we select 50 proteins from the RCSB Protein Data Bank~\cite{berman2003announcing}, with PDB accession codes (PDB ID) listed in Supplementary Section~1.3. Docking simulations are performed using AutoDock Vina~\cite{trott2010autodock,eberhardt2021autodock}, a widely adopted tool known for the accuracy and efficiency in predicting ligand-receptor interactions. This setup enables a direct comparison of the binding behaviors between the original and watermarked molecules, evaluating the impact of \OURS{} on affecting the functionality of molecules. Detailed docking settings and additional results are provided in Supplementary Section~1.3.

Docking performance is assessed using four key metrics, that is binding affinity (Aff), interaction energy between receptor and ligand (INTER), internal ligand energy (INTRA), and root-mean-square deviation (RMSD). Lower values across these metrics indicate more stable docking performance. Fig.~\ref{Fig7} presents the distributions of evaluation metrics for both original and watermarked molecules. Binding affinity is reported as the best binding affinity (Best\_Aff) and average binding affinity (Avg\_Aff). The Best\_Aff values are nearly identical for the original and watermarked molecules, which are -3.86 (0.63, 2.51) kcal/mol and -3.85 (0.64, 2.52) kcal/mol, respectively. The total binding energies (INTER+INTRA) of original and watermarked molecules match precisely at -4.16 (0.81, 3.26) kcal/mol. Meanwhile, the RMSD upper bounds of original molecules and watermarked molecules are comparable, which are 2.46 (0.71, 2.74) \AA~and 2.47 (0.75, 2.96) \AA, respectively. These closely aligned distributions confirm that \OURS{} maintain the docking performance of molecules, preserving molecular suitability for interaction studies.

Representative docking renderings are shown in Fig~\ref{Fig8}, where both the original and watermarked molecules are docked to two proteins (PDB IDs: 7t8r and 8iny). In all cases, the hydrogen-bond geometries and interacting amino-acid residues are nearly identical. For protein 7t8r, the binding affinities of the original and watermarked ligands are -5.80 kcal/mol and -5.79 kcal/mol, respectively. For protein 8iny, the corresponding binding affinities are -5.90 kcal/mol and -5.83 kcal/mol, respectively. These closely matched conformations and binding results further demonstrate that \OURS{} preserves molecular functionality and stability in protein-ligand interactions, supporting the suitability for downstream applications.

\begin{figure*}[htbp]
\centering
\subfigure[]
{
	\begin{minipage}[t]{0.31\linewidth}
    \centering
	\includegraphics[width=1\textwidth]{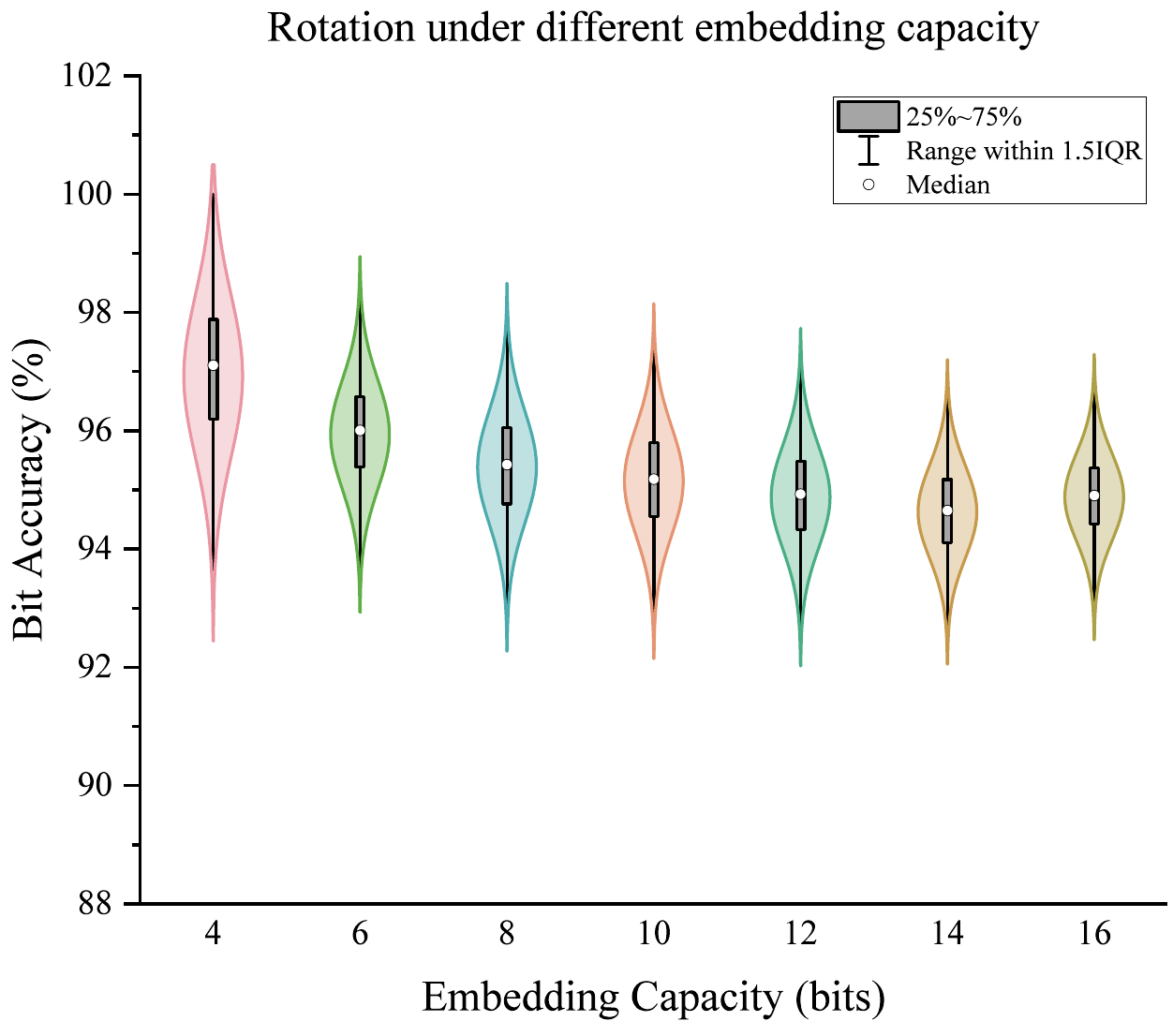}
	\end{minipage}
}
\subfigure[]
{
	\begin{minipage}[t]{0.31\linewidth}
    \centering
	\includegraphics[width=1\textwidth]{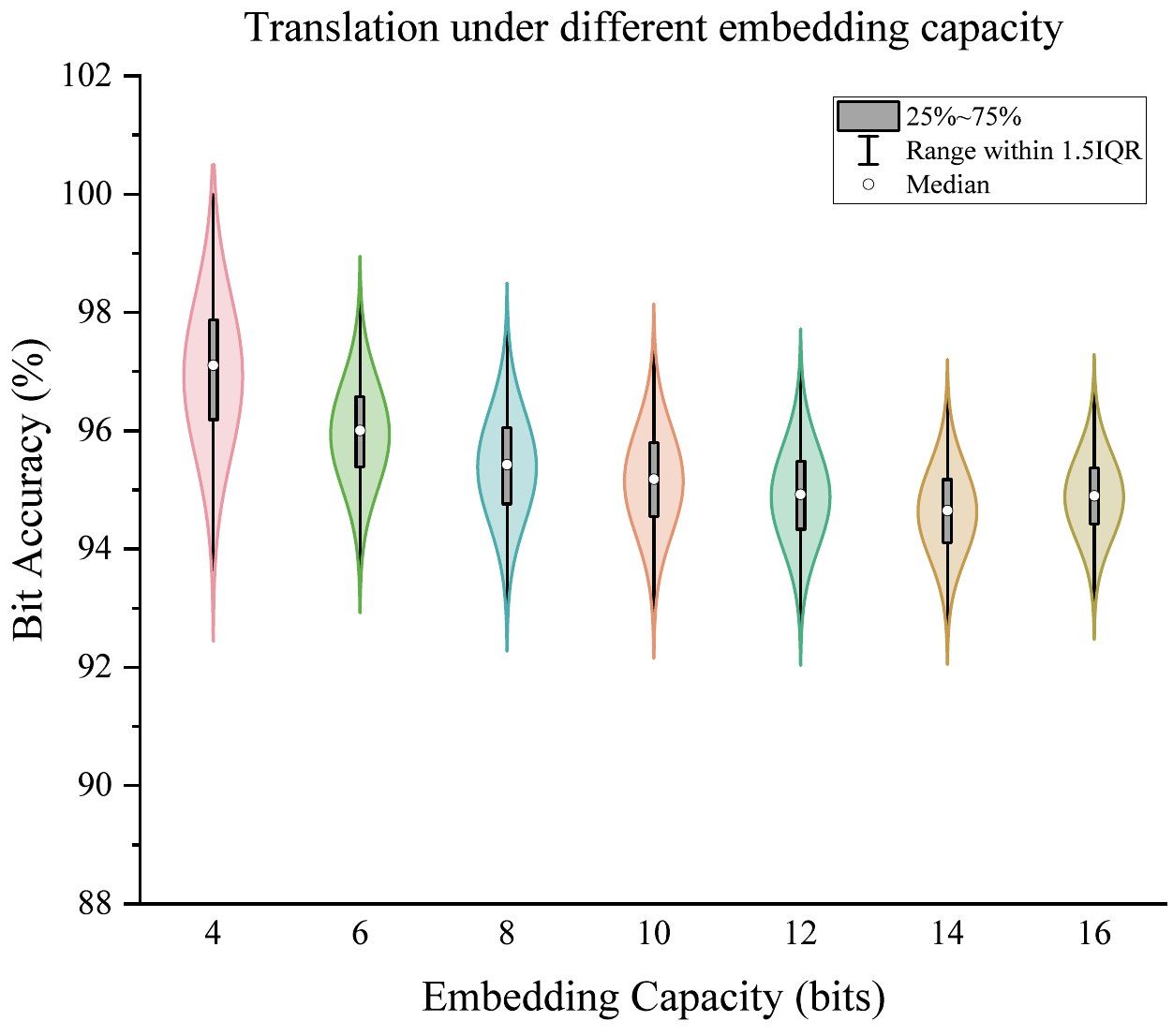}
	\end{minipage}
}
\subfigure[]
{
	\begin{minipage}[t]{0.31\linewidth}
    \centering
	\includegraphics[width=1\textwidth]{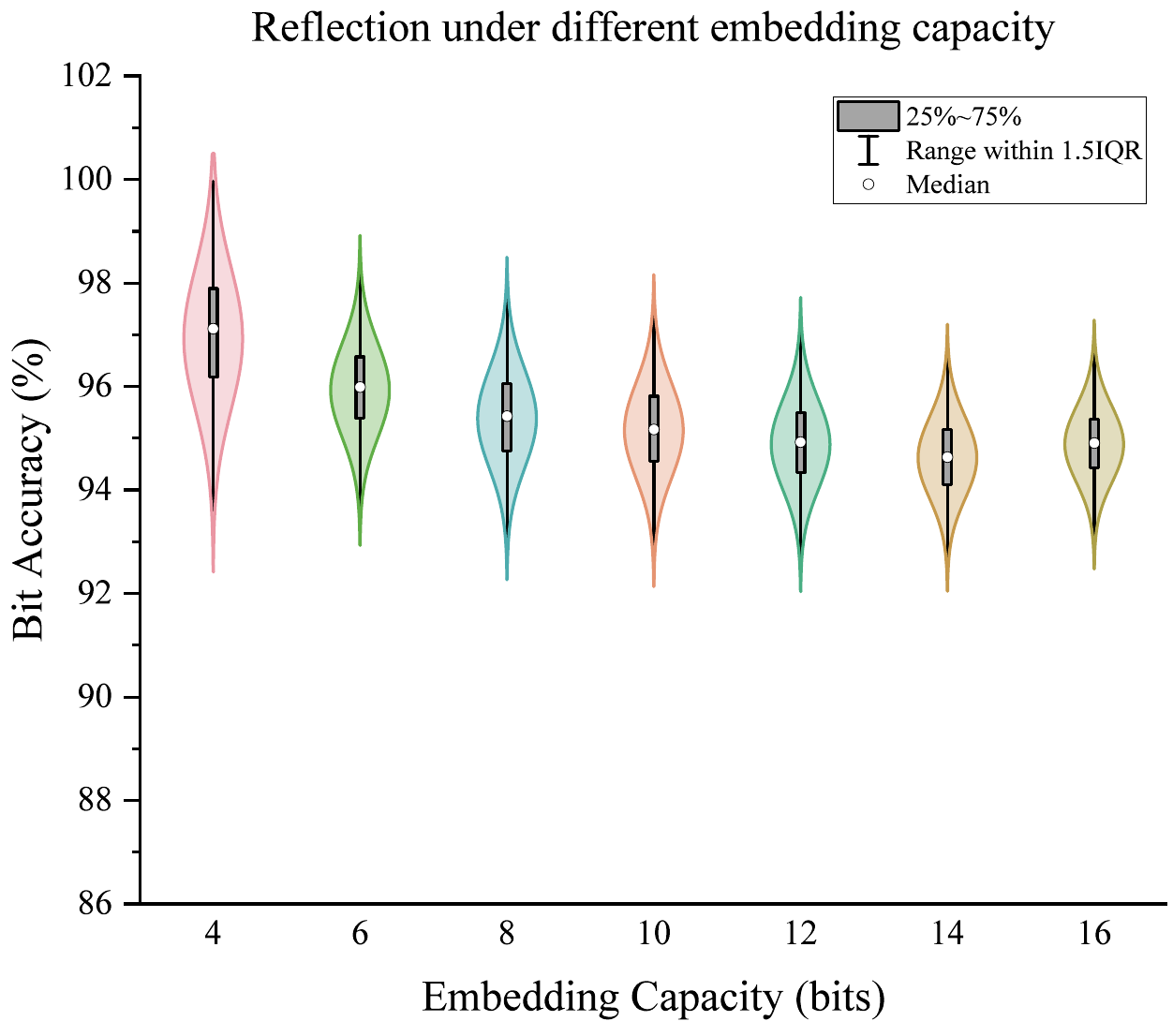}
	\end{minipage}
}
\caption{Bit accuracy of \OURS{} against SE(3) transformations under embedding capacity ranges from 4 bits to 16 bits: (a) robustness against rotation, (b) robustness against translation, and (c) robustness against reflection.}
\label{Fig10}
\end{figure*}

\begin{table*}[htbp]
\caption{Comparison of \OURS{} and three ablation methods (``\OURS{}-w/o AE'', ``\OURS{}-w/o EE'', and ``\OURS{}-w/o AE\&EE'') on the basic properties of molecules from the QM9 and GGEOM-DRUG datasets.}
\centering
\setlength{\tabcolsep}{1.2mm}
\begin{tabular}{l|cccccc|ccc}
\toprule
\multicolumn{1}{c|}{\multirow{2}{*}{Methods}} & \multicolumn{6}{c|}{QM9}                                   & \multicolumn{3}{c}{GEOM-DRUG} \\
\multicolumn{1}{c|}{}                         & Atom Sta & Mol Sta & Validity & Uniq  & Novelty & Bit Acc & Atom Sta & Validity & Bit Acc \\ \midrule
MolMark                                      & 98.78    & 85.52   & 93.43    & 85.31 & 73.29   & 92.91   & 84.90    & 96.07    & 92.18   \\
MolMark-w/o AE                               & 97.30    & 75.10   & 86.56    & 70.41 & 77.52   & 90.78   & 83.94    & 40.52    & 91.12   \\
MolMark-w/o EE                               & 90.54    & 38.64   & 71.04    & 38.43 & 86.02   & 90.83   & 82.44    & 43.64    & 89.89   \\
MolMark-w/o AE\&EE                           & 88.33    & 31.25   & 65.52    & 31.25 & 86.66   & 89.26   & 80.77    & 37.91    & 88.57   \\
\bottomrule
\end{tabular}
\label{Table4}
\end{table*}

\begin{figure*}[htbp]
\centering
\subfigure[]
{
	\begin{minipage}[t]{0.31\linewidth}
    \centering
	\includegraphics[width=1\textwidth]{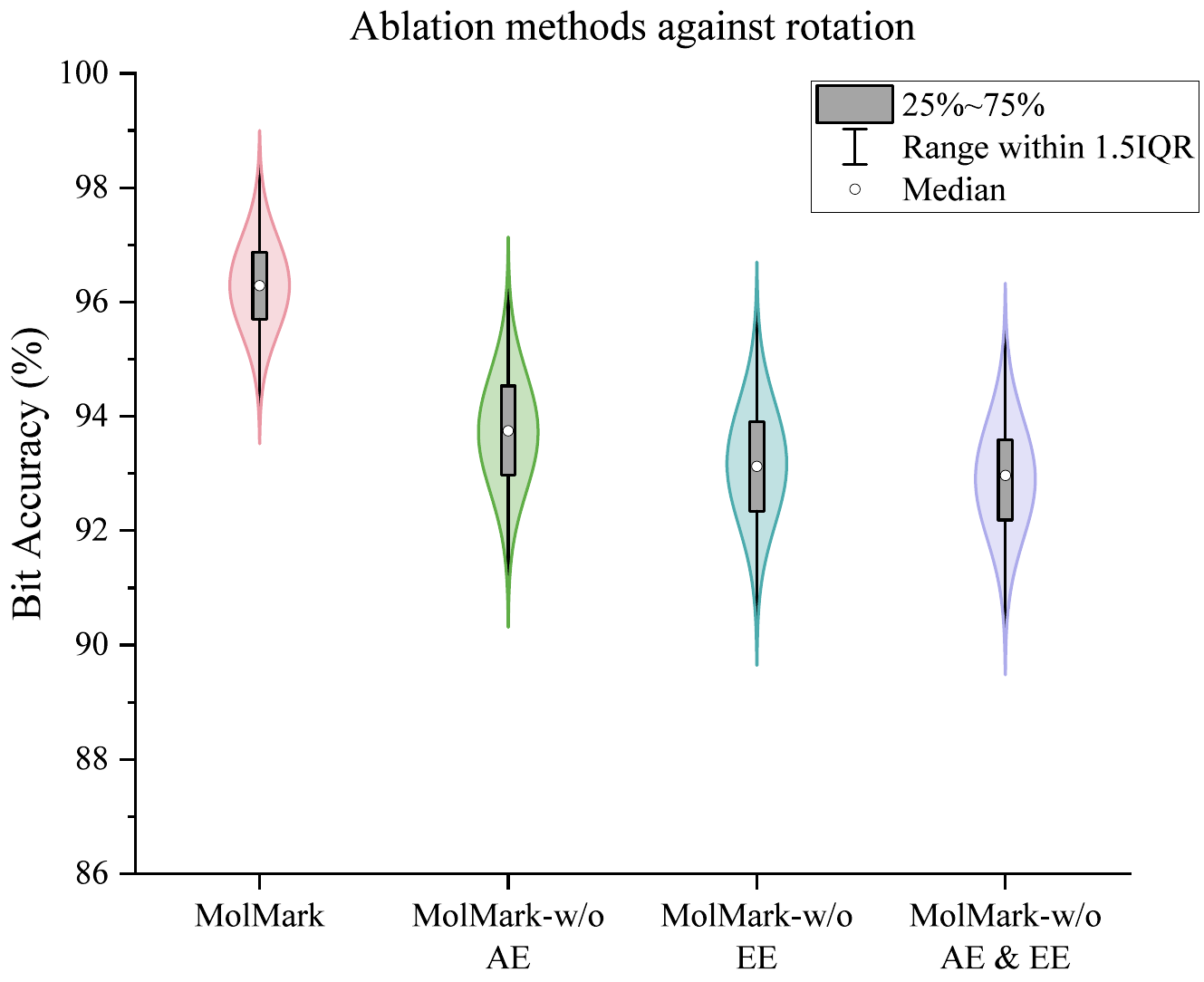}
	\end{minipage}
}
\subfigure[]
{
	\begin{minipage}[t]{0.31\linewidth}
    \centering
	\includegraphics[width=1\textwidth]{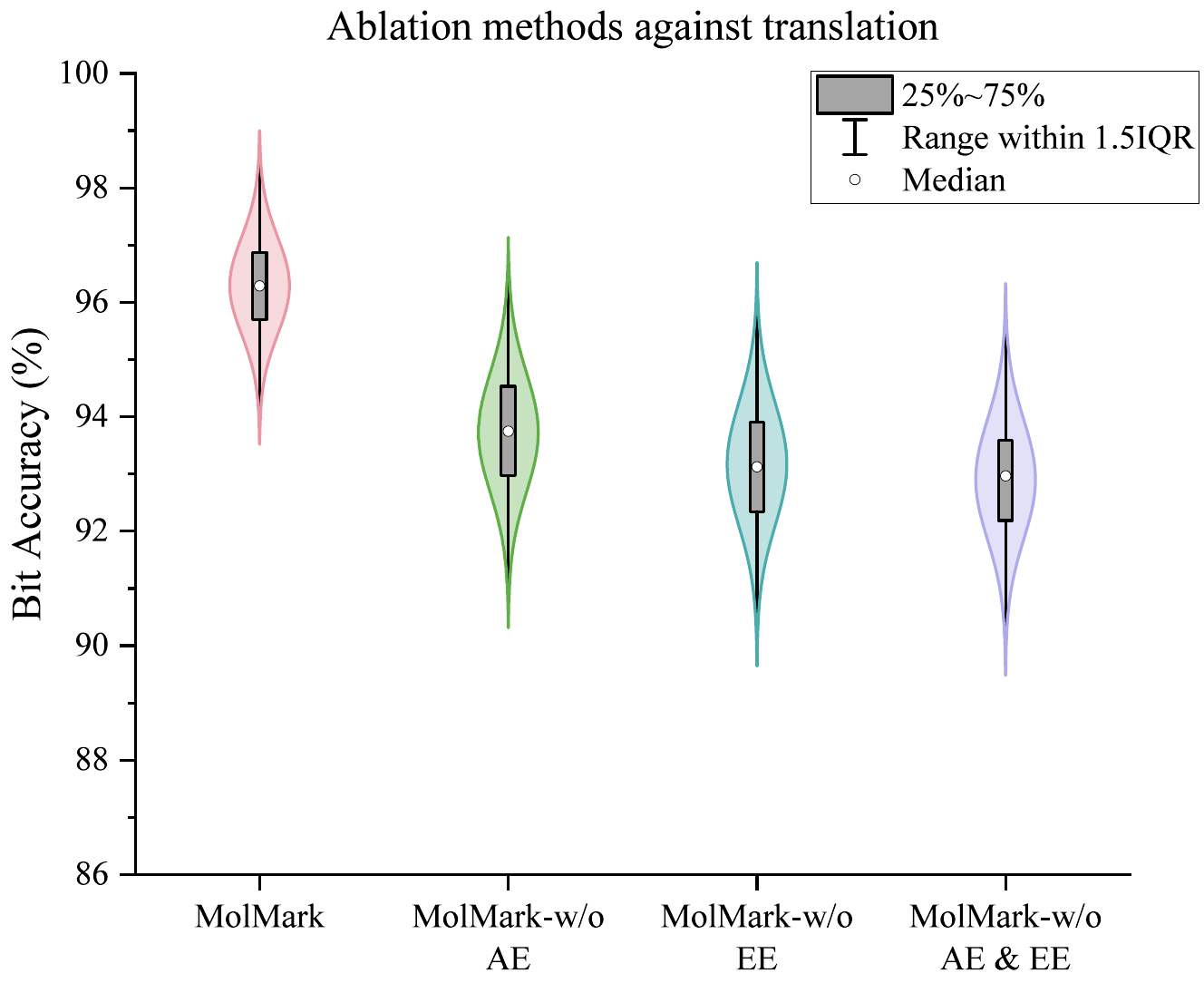}
	\end{minipage}
}
\subfigure[]
{
	\begin{minipage}[t]{0.31\linewidth}
    \centering
	\includegraphics[width=1\textwidth]{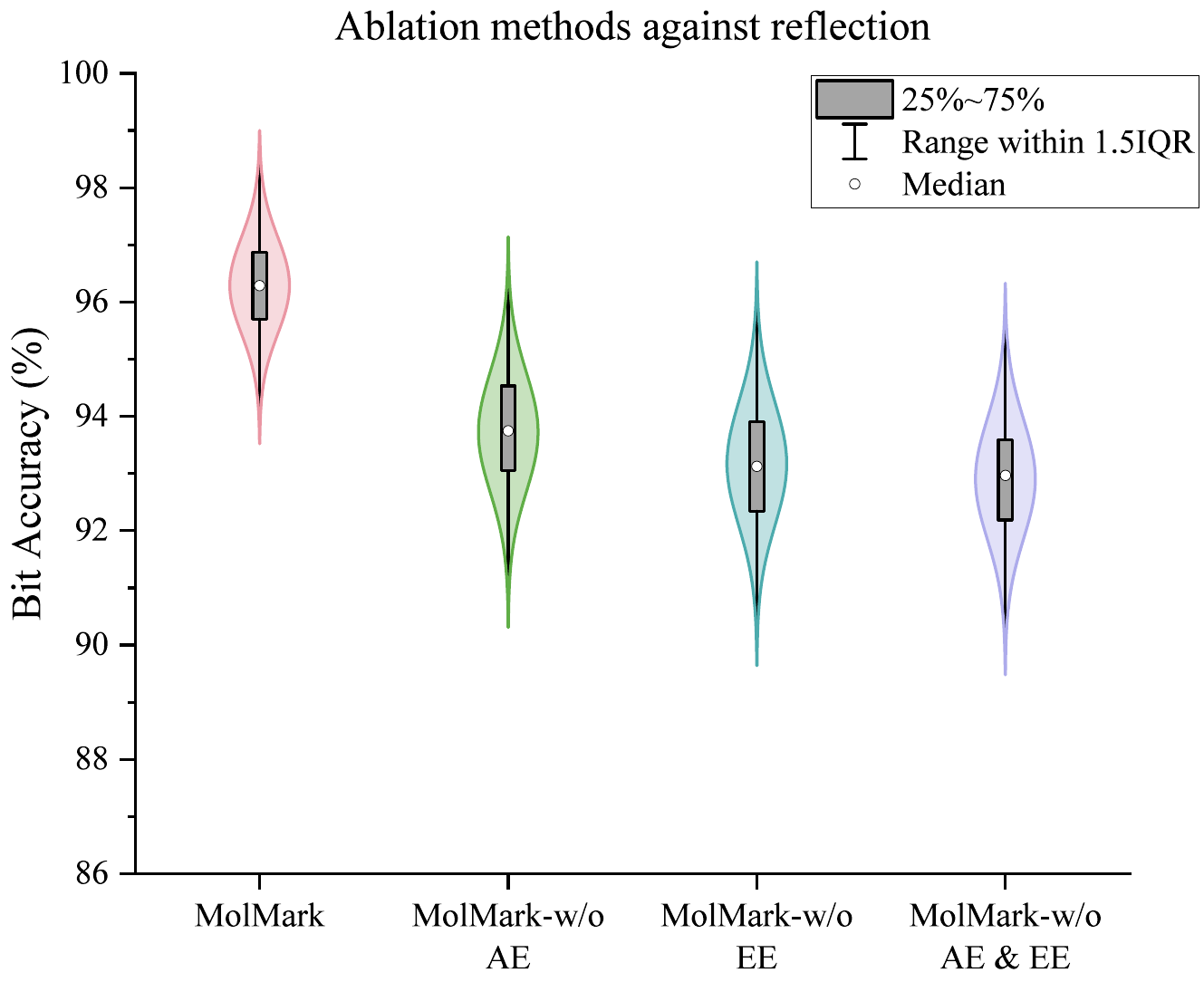}
	\end{minipage}
}
\caption{Bit accuracy of \OURS{} and three ablation methods against SE(3) transformations under embedding capacity ranges from 4 bits to 16 bits: (a) robustness against rotation, (b) robustness against translation, and (c) robustness against reflection.}
\label{Fig11}
\end{figure*}

\subsection{Robustness Against SE(3) Transformations}
\label{section:Robustness}

The robustness of \OURS{} against SE(3) transformations is assessed using bit accuracy as the evaluation metric. Molecules are first generated using GeoBFN, after which \OURS{} embeds watermarks with capacities ranging from 4 to 16 bits. The watermarked molecules are then subjected to SE(3) transformations to emulate practical perturbations in spatial configuration. Following this, the transformed molecules are used to extract the embedded watermarks for calculating the bit accuracy. This measure reflects the robustness and reliability of \OURS{} in supporting molecular copyright protection.

The robustness of \OURS{} against SE(3) transformations is evaluated using bit accuracy as the primary metric. Molecules are first generated with GeoBFN, and \OURS{} embeds watermarks with capacities ranges from 4 bits to 16 bits. The watermarked structures are then subjected to SE(3) transformations to emulate realistic variations in spatial configuration. After transformation, the decoder extracts the embedded watermarks and computes the corresponding bit accuracy. This evaluation quantifies the robustness of \OURS{} to rotation, translation, and reflection, thereby establishing the reliability for molecular copyright protection.

\subsubsection{Robustness against rotation}
\label{Section:Rotation}

Fig.~\ref{Fig9} presents the conformations of transforming a molecule, in which the original molecule is shown in Fig.~\ref{Fig9} (a). Fig.~\ref{Fig9} (b) illustrates the result of rotating a molecule 90 degrees clockwise along the X axis. Although the spatial configuration changes, the physicochemical properties remain unchanged. Fig.~\ref{Fig10} (a) reports the bit accuracies obtained under rotations along the X, Y, and Z axes, where the rotation angle $\alpha$ varies from 0 to 360 degrees in increments of 10 degrees. At an embedding capacity of 8 bits, the bit accuracy reaches 95.43 (1.29, 5.15)\%, and slightly decreases to 94.90 (0.95, 3.82)\% at 16 bits. These results indicate that the embedded watermarks remain reliably extractable under substantial rotational perturbations. Across all embedding capacities and angles, the median bit accuracy exceeds 94.65\%, demonstrating strong robustness against rotation.

Specifically, the bit accuracy of an individual molecule remains unchanged across all rotation angles for each embedding capacity, further confirming the rotational invariance of \OURS{}. Aiming to preserve structural and functional fidelity of watermarked molecules, \OURS{} has made a compromise, that is as embedding capacity increases, the bit accuracy declines. Overall, \OURS{} maintains consistently high performance under diverse rotational transformations. Additional results are provided in Supplementary Section~1.4.1.

\subsubsection{Robustness against translation}

Fig.~\ref{Fig9} (c) shows an example in which a molecule is translated by 0.5 units along the X-axis, resulting in an identical conformation from a visual perspective. To evaluate translation robustness more broadly, molecules are translated along the X, Y, and Z axes with translation values ranging from -1.8 to 1.8 in increments of 0.1. Fig.~\ref{Fig10} (b) presents the distribution of bit accuracies across different embedding capacities. After embedding 8-bit watermark, the bit accuracy of \OURS{} reaches 95.43 (1.29, 5.17)\%, exhibiting only minor fluctuations. Although increasing the embedding capacity leads to reduction in accuracy, all median bit accuracies exceed 94.65\%, indicating strong robustness to translation.

For a fixed embedding capacity, the bit accuracy for a given molecule remains constant across all translation values, confirming the translation invariance of \OURS{}. While constraints imposed to preserve molecular validity introduce minor molecule-specific variations, the overall bit accuracy remains above 95.00\%, supporting the stability and reliability of the method. Supplementary Section~1.4.2 provides additional analysis.

\subsubsection{Robustness against reflection}

Fig.~\ref{Fig9} (d) depicts a reflection of the molecule along the X-axis, which alters the conformation while preserving physicochemical properties. Reflections along the X, Y, and Z axes are applied to evaluate robustness. Fig.~\ref{Fig10} (c) presents the distributions of bit accuracies under different embedding capacities. For an 8-bit watermark, the median accuracy reaches 95.43 (1.31, 5.20)\%, and remains stable across most molecules. When the capacity increases to 16 bits, the accuracy decreases slightly to 94.90 (0.95, 3.75)\%, while maintaining tight central clustering. These results confirm that \OURS{} is robust to reflections across all axes. Additional details are provided in Supplementary Section~1.4.3.

As shown in Fig.~\ref{Fig10}, \OURS{} exhibits strong robustness to SE(3) transformations, including rotation, translation, and reflection. This robustness derives from the discovery of SE(3)-invariant features in Section~\ref{section:Proposed Method}. Leveraging the SE(3)-invariant representations, the embedded watermarks remain stable and recoverable after geometric transformations. Consequently, \OURS{} preserves the structural and physicochemical integrity of molecules while providing a reliable mechanism for copyright protection and traceable molecular identification in downstream applications.

\subsection{Ablation Analysis of Model Components}

\OURS{} is designed with tailored architecture that enables effective molecular copyright protection while preserving the molecular properties. As described in Section~\ref{section:Proposed Method}, the designed model is primarily composed of an atom embedder and an edge embedder. To examine the contributions of different components, we conduct ablation experiment by constructing three variant models: ``\OURS{}-w/o AE'' removes the atom embedder, ``\OURS{}-w/o EE'' removes the edge embedder, and ``\OURS{}-w/o AE\&EE'' removes the atom embedder and the edge embedder. All models are trained on QM9 and GEOM-DRUG datasets with an embedding capacity of 10 bits. Specifically, the basic properties of molecules and the bit accuracy of watermark are adopted as the evaluation metrics in this section.

Molecules generated by GeoBFN are used for comparisons. As shown in Table~\ref{Table4}, \OURS{} consistently yields the smallest deviations in molecular properties, outperforming other variant models. In particular, ``\OURS{}-w/o AE'' achieves better atom stability and molecule stability than ``\OURS{}-w/o EE'', suggesting that the edge embedder plays a more critical role by leveraging interatomic distances to mitigate structural distortions. After removing the atom and edge embedders, ``\OURS{}-w/o AE\&EE'' demonstrates the largest property deviations, highlighting the complementary nature of the two embedders.

Robustness to SE(3) transformations is assessed by applying rotation, translation, and reflection to the watermarked molecules. Fig.~\ref{Fig11} shows the distributions of bit accuracies under the SE(3) transformations. Based on the results, we can find that \OURS{} achieves the highest median accuracy, followed by ``\OURS{}-w/o AE'', ``\OURS{}-w/o EE'', and ``\OURS{}-w/o AE\&EE''. For example, under the attack of rotation, the median bit accuracies of the three variants are 93.75\%, 93.13\%, and 92.97\%, respectively. Despite these differences, all models maintain median accuracies above 92.97\% across the SE(3) transformations, demonstrating the effectiveness of SE(3)-invariant features.

Overall, the ablation results underscore the importance of jointly leveraging atom-level and edge-level geometric cues. The atom embedder and edge embedder substantially maintains the properties of molecules and strengthens robustness under SE(3) perturbations. These findings suggest that architectural refinements are important for advancing molecular copyright protection, motivating further investigation into exquisite model designs.

\section{Conclusion}
\label{section:Conclusion}
In this paper, we propose \OURS{}, the first watermarking framework tailored to AI-generated molecular structures. Previous watermarking methods fail to protect molecules that destruct the basic properties and downstream functionalities. Considering the compactness of molecular structures and the strict chemical bonding constraints, we exquisitely design \OURS{}, which embeds digital signatures into molecules while preserving the structural and functional integrity. The framework of \OURS{} is composed of a watermark encoder and decoder. The encoder learns to embed watermarks by minimally adjusting atom positions, in which the atom and edge features are fully utilized to preserve the chemical validity and stability. Meanwhile, the decoder discovers SE(3)-invariant representations to enable accurate extraction after common visualization/manipulation operations. In this way, \OURS{} achieves high watermark fidelity with minimal perturbation to molecular properties.

Extensive experiments are conducted on the QM9 and GEOM-DRUG datasets and two unconditional generative models GeoBFN and GeoLDM. We examine the feasibility of embedding watermarks into molecular structures, the influence of watermark capacity on molecular properties, comparisons against previous watermarking baseline methods, the impact on physicochemical properties, the functional behavior in molecular docking, and the robustness against geometric transformations. Across all experiments, \OURS{} achieves consistently high bit accuracy ($\ge$95\% with 16-bit watermarks) while preserving performance comparable to original molecules. In contrast, previous watermarking methods reduce molecular stability to nearly 0\%, whereas \OURS{} maintains atom-level and molecule-level stability above 94\%.

In summary, \OURS{} provides the first step toward trustworthy and accountable protection for AI-driven molecular design. By enabling verifiable attribution without compromising function, \OURS{} supports open innovation while safeguarding copyrights. As generative models become increasingly embedded in drug discovery and materials design, frameworks like \OURS{} will be indispensable for ensuring the innovative and accountable of scientific ecosystem.


\section*{Code availability}
The source codes are available via GitHub at \textcolor{blue}{\url{https://github.com/RunwenHu/MolMark}}. The demo of \OURS{} is publicly available at \textcolor{blue}{\url{https://molmarkdemo.streamlit.app}} and the project page is available at \textcolor{blue}{\url{https://molmark1.github.io}}.

%

%

\ifCLASSOPTIONcompsoc
  \section*{Acknowledgments}
\else
  \section*{Acknowledgment}
\fi

This work was supported in part by the National Natural Science Foundation of China under Grant.

\ifCLASSOPTIONcaptionsoff
  \newpage
\fi

\bibliographystyle{IEEEtran}
\bibliography{reference}

\vspace{-1cm}
\begin{IEEEbiography}[{\includegraphics[width=1in, height=1.20in, clip, keepaspectratio]{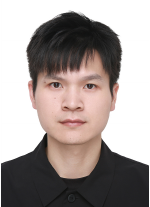}}]
{Runwen Hu} received the B.S. degree and the Ph.D. degree from the College of Information Science and Technology (2019) and the College of Cyber Security (2024), Jinan University, Guangzhou, China. He is currently a postdoctoral fellow at City University of Hong Kong. His current research interests include reversible data hiding and robust data hiding.
\end{IEEEbiography}
\begin{IEEEbiography}[{\includegraphics[width=1in, height=1.20in, clip, keepaspectratio]{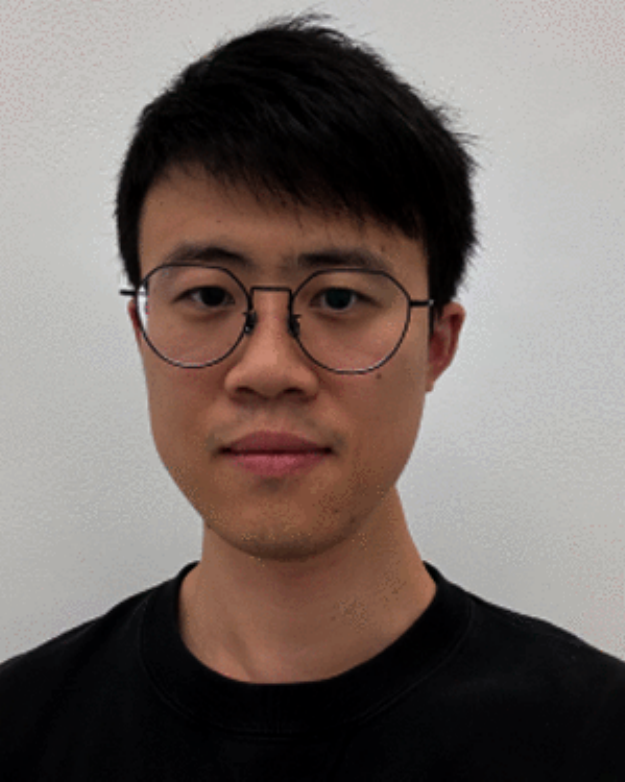}}]
{Peilin Chen} received the B.E. degree in software engineering from Sun Yat-sen University, Guangzhou, China, in 2018, and the Ph.D. degree in computer science from the City University of Hong Kong (CityUHK), Hong Kong, in 2023. He is currently a Postdoctoral Researcher with the Department of Computer Science, CityUHK. His research interests include visual data processing, semantic communication, and multimodal large language models.(Based on document published on 9 September 2025).
\end{IEEEbiography}
\begin{IEEEbiography}[{\includegraphics[width=1in, height=1.20in, clip, keepaspectratio]{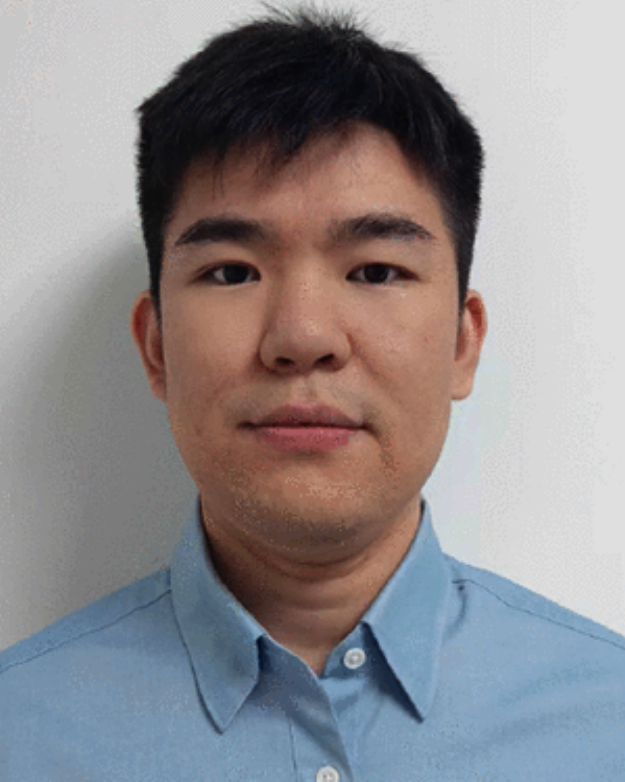}}]
{Keyan Ding} received the B.E. degree from China Jiliang University, Hangzhou, China, in 2015, the M.S. degree from Soochow University, Suzhou, China, in 2018, and the Ph.D. degree from the Department of Computer Science, City University of Hong Kong, Hong Kong, in 2022. He is currently a Researcher with the ZJU-Hangzhou Global Scientific and Technological Innovation Center, Zhejiang University, Hangzhou, China. His research interests include computer vision and image processing
\end{IEEEbiography}
\vspace{-1cm}
\begin{IEEEbiography}[{\includegraphics[width=1in, height=1.20in, clip, keepaspectratio]{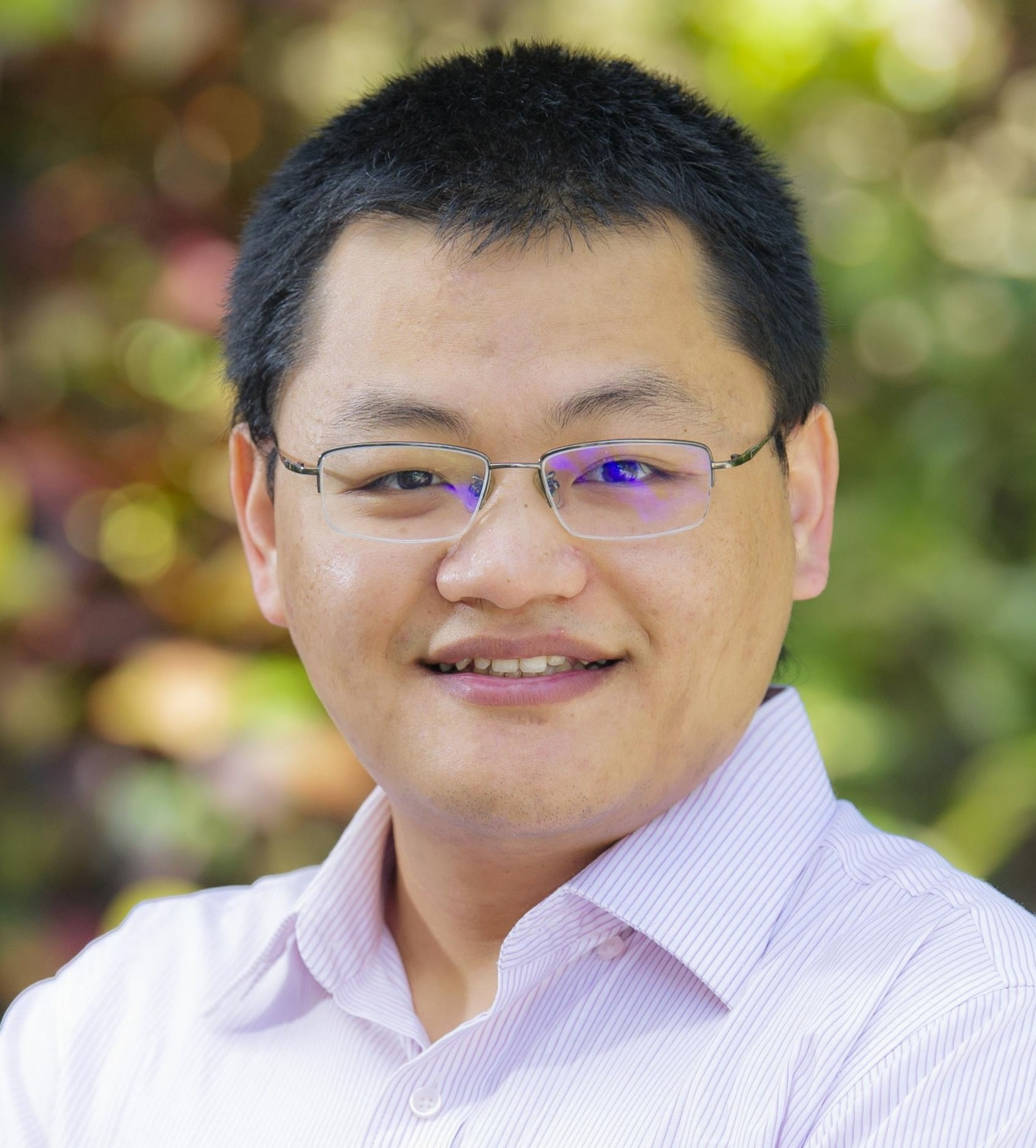}}]
{Shiqi Wang} (Senior Member, IEEE) received the B.S. degree in computer science from Harbin Institute of Technology, in 2008, and the Ph.D. degree in computer application technology from Peking University, in 2014. From 2014 to 2016, he was a Post-Doctoral Fellow with the Department of Electrical and Computer Engineering, University of Waterloo, Waterloo, ON, Canada. From 2016 to 2017, he was a Research Fellow with the Rapid-Rich Object Search Laboratory, Nanyang Technological University, Singapore. He is currently an Associate Professor with the Department of Computer Science, City University of Hong Kong. He has proposed more than 50 technical proposals to ISO/MPEG, ITU-T, and AVS standards, and authored or co-authored more than 200 refereed journal articles/conference papers. His research interests include video compression, image/video quality assessment, and image/video search and analysis. He is also the Technical Program Co-Chair of IEEE ICME 2024. He received the Best Paper Award from IEEE VCIP 2019, ICME 2019, IEEE Multimedia 2018, and PCM 2017. His co-authored article received the Best Student Paper Award in the IEEE ICIP 2018. He was a recipient of the 2021 IEEE Multimedia Rising Star Award in ICME 2021. He served or serves as an Associate Editor for IEEE Transactions on Circuits and Systems for Video Technology, IEEE Transactions on Multimedia, IEEE Transactions on Image Processing, and IEEE Transactions on Cybernetics.
\end{IEEEbiography}

\end{document}